\documentclass[preprint,12pt]{elsarticle}
\usepackage[a4paper, total={6in, 9in}]{geometry}

\usepackage{amssymb}
\usepackage{graphicx}
\usepackage{amsmath}
\usepackage{url}
\usepackage{hyperref}
\hypersetup{
     colorlinks   = true,
     citecolor    = blue
}

\usepackage{longtable}
\usepackage{adjustbox}	% - Compressing tables
\usepackage{arydshln}
\usepackage{float}
\usepackage{multirow}
\usepackage[caption=false,font=footnotesize]{subfig}

\journal{Expert Systems}

\graphicspath{ {./images/} }

\begin{document}
\begin{frontmatter}

\title{Multi-Label Logo Recognition and Retrieval based on Weighted Fusion of Neural Features}

% Multi-Label Logo Classification and Similarity Search using Convolutional Neural Networks

\author[1]{Marisa Bernabeu}
\ead{mbernabeu@dlsi.ua.es}
\author[1]{Antonio Javier Gallego\corref{cor1}}
\ead{jgallego@dlsi.ua.es}
\author[1]{Antonio Pertusa}
\ead{pertusa@ua.es}
\address[1]{University Institute for Computing Research, University of Alicante, Carretera San Vicente del Raspeig s/n, 03690 San Vicente del Raspeig, Alicante, Spain}
\cortext[cor1]{Corresponding author. Tel.: (+34) 965903400 ext. 2038}

%\author{Marisa Bernabeu \and Antonio Javier Gallego\orcidID{0000-0003-3148-6886} \and
%Antonio Pertusa\orcidID{0000-0002-9445-5529}
%}
%

% LIMIT: 250 words
\begin{abstract}
Classifying logo images is a challenging task as they contain elements such as text or shapes that can represent anything from known objects to abstract shapes. While the current state of the art for logo classification addresses the problem as a multi-class task focusing on a single characteristic, logos can have several simultaneous labels, such as different colors. This work proposes a method that allows visually similar logos to be classified and searched from a set of data according to their shape, color, commercial sector, semantics, general characteristics, or a combination of features selected by the user. Unlike previous approaches, the proposal employs a series of multi-label deep neural networks specialized in specific attributes and combines the obtained features to perform the similarity search. To delve into the classification system, different existing logo topologies are compared and some of their problems are analyzed, such as the incomplete labeling that trademark registration databases usually contain. The proposal is evaluated considering 76,000 logos (7 times more than previous approaches) from the European Union Trademarks dataset, which is organized hierarchically using the Vienna ontology. Overall, experimentation attains reliable quantitative and qualitative results, reducing the normalized average rank error of the state-of-the-art from 0.040 to 0.018 for the Trademark Image Retrieval task. Finally, given that the semantics of logos can often be subjective, graphic design students and professionals were surveyed. Results show that the proposed methodology provides better labeling than a human expert operator, improving the label ranking average precision from 0.53 to 0.68.
\end{abstract}

%Dataset
% 76,000 (9 años)
% Previous approach: 11,000 and 1 year
% The use of a much larger corpus (logos from the European Union Intellectual Property Office, EUIPO, from nine years rather than only one year)

%  NAR 0.040   - 0.018 Normalized Average Rank (NAR)

% Results obtained in the survey of design students and professionals using the LRAP metric, compared with the result obtained by our proposal. Higher LRAP values indicate better results.}
%\textbf{Average}        & 0.5292		& 0.6833
% Improvement: 0.1541


\end{frontmatter}

% ---------------------------------------------------------------------------
\section{Introduction}

The detection and recognition of logos is an important task given that companies need to detect the use of their logos in images \citep{Bianco_2017}, social media \citep{10.1007/978-3-030-20518-8_11} and sports events \citep{Kostinger_planartrademark}, or to discover unauthorized usages and plagiarism. Moreover, to register trademarks, it is necessary to verify that there are no similar logos within the same business sector. This is a relevant issue owing to the volume of applications for trademark registration and the size of the databases containing existing trademarks, moreover when considering how costly it would be for humans to make these comparisons visually~\citep{Perez2018}. 

Most previous computer vision approaches for logos have focused on the Trademark Image Retrieval (TIR) task, which consists of performing a similarity search to obtain the most similar logos given a query image. Schietse et al.~\cite{Schietse2007} describe the main challenges that TIR systems confront. Logo images differ from real pictures since they are artificially created and designed to have a visual impact. In addition, they may contain only text, images, or a combination of both. The most relevant feature for humans to characterize a logo is probably the shape. However, automatic shape classification is a challenging task. In addition to the structure of the elements that comprise a logo and its organization, semantic interpretation must also be considered to determine the objects present in logos. This is a very complex task related to how humans perceive and interpret images.

Color also plays an essential role in designing and characterizing a logo. Brands within the same business area often use similar colors owing to their cultural and social connotations. However, this is not always the case since organizations may also use color to differentiate themselves from the competition. For example, the authors of \cite{Logos01} describe the case of the technology company Gear6, which uses the color green to distinguish itself from its competitors. Color is, therefore, important when performing TIR, but it is also necessary to consider that logos sometimes lose their color and that we can also find versions in grayscale. 

\begin{figure}[t]
\centering
\includegraphics[width=0.9\textwidth]{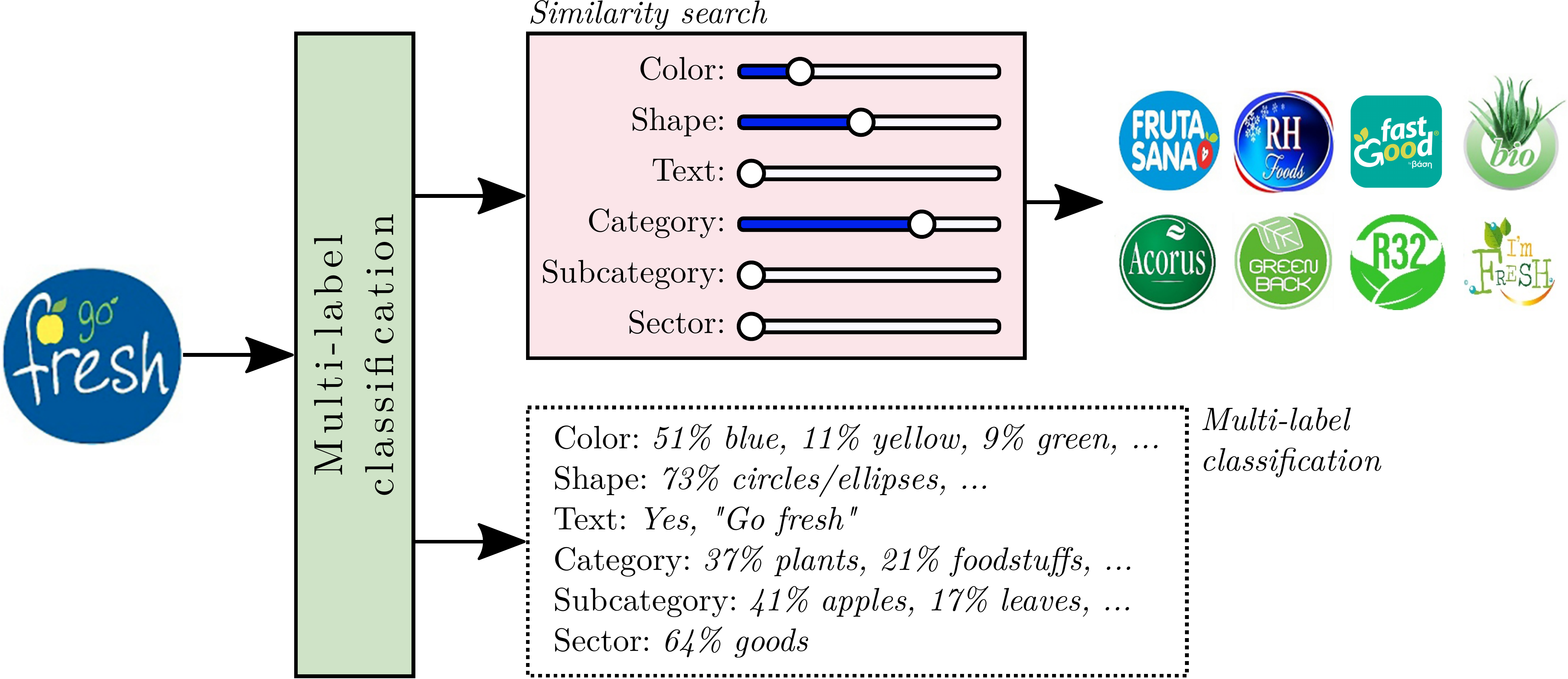}
\caption{Overview of the proposed method.}
\label{fig:app}
\end{figure}

This work presents a method with a twofold purpose. In addition to retrieving the most similar logos according to criteria provided by a user, it also allows performing multi-label classification. Figure \ref{fig:app} shows an overview of the method. First, a multi-label architecture is proposed to classify different features of the logo, such as color, shape, and figurative elements (semantics). This stage aims to facilitate the labeling of brands since the output contains a series of label options with an associated probability to assist the operator in this process, as seen in the bottom white box from the figure. A similarity search module is also added (top box), allowing the operator to adjust the search criteria. This module can be used to find similar designs or detect plagiarism.

%estudio preliminar en ibprialogos
A preliminary study on multi-label logo classification and similarity search was proposed in~\cite{ibpriaLogos}. We extend this method and its evaluation by making the following contributions:
%The work presented builds upon the approach proposed in \cite{ibpriaLogos} and extends this method and its evaluation by making the following contributions:

\begin{enumerate}
\item The addition of a preprocessing stage for text detection, which includes an inpainting method to improve the retrieval results. 

\item The inclusion of additional neural models and changes to the final stage of the similarity search using an alternative method that improves the results.  

\item The analysis of existing topologies to understand the hierarchical Vienna classification system used for comparing and retrieving logos from a design perspective. 

\item The proposal of multi-label tagging by grouping Vienna classification codes.

\item The use of a much larger corpus (logos from the European Union Intellectual Property Office, EUIPO, from nine years rather than only one year) to train the networks.

\item A comparison of the proposal with 17 state-of-the-art methods, outperforming their results.

\item The presentation of a qualitative evaluation through surveys of graphic design students and expert designers.
\end{enumerate}

The remainder of the paper is structured as follows: the background related to the topic is introduced in Section~\ref{sec:background}. Then, the proposed approach is developed thoroughly in Section~\ref{sec:method}. Next, section~\ref{sec:setup} describes the experimental setup considered, while the results obtained and an analysis of them are included in Section~\ref{sec:evaluation}. Finally, the general conclusions are discussed in Section~\ref{sec:conclusions}.

% ---------------------------------------------------------------------------
\section{Background}
\label{sec:background}

The state of the art of methods used for TIR and multi-label retrieval are reviewed in the first part of this section, after which the available datasets and the topologies used for their classification are detailed.

\subsection{Trademark Image Retrieval}

Most traditional methods have addressed TIR by extracting a series of handcrafted features and using them to feed a $k$-Nearest Neighbor ($k$NN)~\cite{DudaHS01} to obtain a ranking of the most similar logos. Some of the features used for this comparison include methods based on color histograms~\citep{Ghosh2015}, shape~\citep{ShapeTrademark10}, local descriptors such as SIFT~\citep{chiambrand}, or a combination of them~\citep{KUMAR2016370,Guru18}. In some cases, the dimensionality of these features is reduced with Bags of Words \cite{IandolaSGK15}. In addition, distance metrics are generally employed for comparing the features processed, although more complex approaches based on template matching have also been proposed~\cite{Pornpanomchai2015}.

Handcrafted features for this task are also found in more recent works, such as \cite{lourenco2019hierarchyofvisualwords}, which introduces HoVW (Hierarchy of Visual Words). This TIR method decomposes images into simpler geometric shapes and defines a descriptor for binary logo image representation by encoding the hierarchical arrangement of component shapes. Nonetheless, most current TIR methods use deep learning \cite{lecun2015deep} architectures. For example, in \cite{chiambrand} an AlexNet network \citep{Krizhevsky2012} with a sliding window is used to find logos in real images. The authors of \cite{IandolaSGK15} evaluated GoogleNet-based Convolutional Neural Networks (CNN) architectures for the brand classification of logos. More recently, \cite{Perez2018} proposed a combination of descriptors extracted from a VGG-19 network to find similarities using the cosine distance, and \cite{Xia2019} used Transform-invariant Deep hashing for TIR by learning transformation-invariant features. 

Our proposal is also based on deep learning. However, it employs a much more versatile method that combines the descriptors learned by a set of multi-label networks specialized in the classification of the different characteristics of logos. 
Most TIR methods reviewed rely on the brand to perform the similarity search or classification. However, a brand's image may change over time, in addition to the fact that the generic comparison of a logo does not allow its classification. For this, it is essential to consider distinctive characteristics of the logo, such as the use of colors, the semantic meaning of shapes, etc.
The proposed approach makes it possible to perform similarity search based on different criteria while simultaneously taking advantage of these characteristics to perform multi-label logo retrieval.

%------------------------
\subsection{Multi-label logo image retrieval}

As shown in the previous section, many TIR works exist in the literature. However, only a few approaches aim to classify logos using features other than brands. In this case, samples may have more than one simultaneous label (for example, several colors, shapes, or figurative elements annotated for the same logo), making this problem a Multi-Label Classification (MLC) task. MLC differs from the traditional multi-class classification in that labels are treated as independent target variables (i.e., not relying on their mutually exclusive assumption). This is suboptimal for MLC because the dependencies among classes cannot be leveraged. 

MLC has received significant attention in recent machine learning literature owing to its interesting applications. During the past decade, great strides have been made in this emerging paradigm. In \cite{ZhangZhou2014}, there is a review on this area emphasizing state-of-the-art multi-label learning algorithms. Multi-label methods are used in applications as diverse as text categorization \citep{MLCtext}, music categorization \citep{MLCmusic}, or semantic scene classification \citep{MLCscene}. However, to the best of our knowledge, the literature contains no examples of MLC-related works applied to logos, except for \cite{ibpriaLogos}. As argued in the introduction section, features such as color, shape, or semantic meaning play a key role in logo classification. Given the particular characteristics of this task, it is of particular interest to developing multi-label systems that make it possible to classify and search for logos based on different criteria that the user can configure.

This paper proposes an MLC approach applied to logos that, in addition to extending the methodology proposed in \cite{ibpriaLogos} and obtaining better results, also broadens the set of labels considered and considerably expands both the experiments and the analysis of the results obtained.

\subsection{Datasets and topologies}
\label{sec:topologies}

Reliable image datasets are crucial for tackling this task, although the corpora used in former works are not generally publicly available for copyright reasons~\citep{Ghosh2015,Kalantidis2011}. As a result, it has not been until recently that some free logo datasets appeared. Some examples are the Large Logo Dataset (LLD) \citep{Sage_2018}, which consists of more than 600,000 logos obtained from the Internet, METU \citep{Tursun2015METU,Tursun2017}, which contains 923,343 trademark images, and Logos in the Wild~\citep{Tzk2018OpenSL}, in which 11,054 images are labeled within 871 brands. 

All the datasets mentioned above, and most of those used by the state-of-the-art methods, are labeled only by brand, as it is assumed that logos from the same brand tend to be similar. However, brands may evolve different versions of their logos (e.g., Disney has changed its logo more than 30 times~\citep{IandolaSGK15}). These differences may include changes in the background, color, texture, or shape, thus making the different versions of a logo very different in appearance and signifying that relying on visual similarity is not always a suitable means to classify logos from the same brand. 

It is, therefore, complicated to establish a categorization method for logos. One of the topologies accepted as a standard by the different agencies for trademark registration worldwide is the Vienna classification, which was developed by the World Intellectual Property Organization (WIPO) \cite{world2002international}. It is used by the European Union Intellectual Property Office (EUIPO) and the United States Patent and Trademark Office (UPSTO), among others, to classify their datasets.

Vienna classification (which will be described in detail in Section~\ref{sec:vienna}) is an international system used to label different characteristics of trademarks by employing a hierarchical topology ordered from the most general to the most specific concepts. It allows images to be labeled with metadata indicating their figurative meaning (semantics), color, shape, and whether or not they contain text. Several patent and trademark agencies have recently released datasets along with their metadata, thus making possible works such as \cite{Rusinol2011} or \cite{ibpriaLogos}, which use this information for the classification or comparison of logos. 

In addition to this labeling, there are classifications that professionals frequently use, such as the topologies proposed by Wheeler \cite{wheeler2013designing} and Chaves \cite{chaves2003marca}, which are based on other kinds of criteria. 

Wheeler classifies logos into three general categories:  Wordmarks (a freestanding acronym, company name, or product name), emblems (logos in which the company name is inextricably connected to a pictorial element), and only symbols (which are subdivided into letterforms, pictorial marks, and abstract/symbolic marks). However, the boundaries among these categories are pliant, and many logos may combine elements from more than one category. 

The alternative categorization proposed by Chaves \cite{chaves2003marca} is similar but is more detailed and based on formal aspects. In this case, there are four main categories: logotypes (which is equivalent to Wheeler’s ``Wordmarks'' category, but with three subtypes: pure logotype, logotype with background, and logotype with accessory), logo-symbol (equivalent to emblems), logotypes with symbols, and only symbols, which, as occurs in Wheeler’s version, is divided into three subtypes. 

Given the relevance of the labeling method for the proposed methodology, the following section describes the Vienna classification and its relationship with the Wheeler and Chaves topologies that are oriented to designers.

% ---------------------------------------------------------------------------
\subsubsection{Vienna classification}
\label{sec:vienna}

The Vienna classification \cite{world2002international} proposes a hierarchical topology of logos, in which each image can be labeled with a series of codes indicating its semantics, shape, color, etc. It defines a set of 29 main categories, which are in turn divided into 2\textsuperscript{nd} and 3\textsuperscript{rd} level categories, creating a classification with hundreds of possible labels. The complete list of top-level categories can be seen at \ref{app:vienna}. Each code is indicated using the XX.YY.ZZ pattern. For example, the 5.9.1 code would assign the tag ``carrots'' to a logo. The hierarchy of this code indicates that it belongs to the category of 2\textsuperscript{nd} level 5.9 ``vegetables'' and the main category 5 ``plants''.

This hierarchical organization makes it possible to group logos by different levels of labels and use higher hierarchical levels when the 3\textsuperscript{rd} or 2\textsuperscript{nd} levels have too much detail, are not very representative, contain few samples, or are ambiguous. 

It is also necessary to consider that the labeling from trademark agencies is usually not exhaustive since only the most distinctive characteristics of brands are typically annotated. This means that incomplete or contradictory labeling can sometimes be found (e.g., a logo that has three colors and only one of them is labeled).

In this work, we propose solving some problems by grouping these codes according to their characteristics and semantics. The intention is not to replace Vienna but to carry out a selection of labels to keep those most useful for their application in machine learning tasks. The following four categories are, therefore, defined, in which the Vienna codes that uniquely describe these characteristics are selected or grouped:

\begin{itemize}
    \item \textbf{Figurative}. This includes the Vienna codes from 1 to 25, which indicate categories related to the figurative or semantic meaning of the logo. For this category, we differentiate between the 1\textsuperscript{st} and 2\textsuperscript{nd} levels of the Vienna hierarchy, which we respectively denominate as the main category (which contains the 25 codes from the 1\textsuperscript{st} level) and the sub-category (with 123 possible classes).
    
    \item \textbf{Colors}. Vienna category 29 refers to colors, although many codes indicate their number (e.g., 29.01.12 means that there are two predominant colors). It is, therefore, proposed that they should not be considered since they do not provide relevant information. After performing this filtering, the set of selected color codes is reduced to 13 (included in \ref{app:vienna}).
    
    \item \textbf{Shapes}. In category 26, different types of shapes are labeled, including circles, triangles, quadrilaterals, etc. In this case, the 3\textsuperscript{rd} level of labeling is very specific and sometimes ambiguous (e.g., curved lines versus wavy lines, or dotted lines versus broken lines). We, therefore, propose using only up to the 2\textsuperscript{nd} level. Moreover, codes 26.07 and 26.13 are grouped in category 26.5 (Other polygons) since a defined shape is not visually identified. After this grouping, a list of 7 possible shape categories was eventually obtained (see \ref{app:vienna}).
    
    \item \textbf{Text}. Category 27 defines the text and its characteristics. This category is also too detailed (e.g., there are 20 different codes to indicate the appearance or shape of the text and as many to indicate the style of the font). Since the specific text in the logo is often made up of acronyms, monograms, or brand names that do not contribute much to the calculation of the similarity between logos, we, therefore, propose to label only the presence or absence of text in the image.

\end{itemize}

Table \ref{tab:classification_equiv} shows a summary of the equivalence among the topologies proposed by Wheeler and Chaves and their relation to the proposed Vienna code groups. These equivalences allow us to determine the most relevant characteristics of logos when preparing or analyzing their design. For example, color and shape are features that appear in all types of designs and can, therefore, help the most to distinguish them. This is not the case with the presence of text or figurative elements, although they are very useful in determining some of the logo features. In summary, there is a relationship among the different topologies, signifying that the Vienna codes can describe the remaining classifications.

\begin{table}[ht]
\renewcommand{\arraystretch}{1.2}
\setlength{\tabcolsep}{6pt}
\caption{Relationship between the Vienna classification and the topologies proposed by Wheeler and Chaves.}
\label{tab:classification_equiv}
\centering
\begin{adjustbox}{width=\columnwidth, keepaspectratio}
\begin{tabular}{lllcccc}
\hline
&&& \multicolumn{4}{c}{\textbf{Vienna}}   \\ 
& \textbf{Wheeler}                  & \textbf{Chaves}                   & \rotatebox[origin=l]{90}{\textbf{Figurative}}
& \rotatebox[origin=l]{90}{\textbf{Color}}
& \rotatebox[origin=l]{90}{\textbf{Shape}}
& \rotatebox[origin=l]{90}{\textbf{Text}}  \\ \hline

& \textbf{Wordmark}                   & \textbf{Logotype:}                  &                                                       \\  
\textbf{Nominal}        && $\diamond$ Logotype with background & --         & \checkmark & \checkmark & \checkmark     \\  \cdashline{3-3}  \cdashline{4-7}        % Shape
\textbf{Identifier}     && $\diamond$ Pure logotype            & --         & \checkmark & --         & \checkmark     \\   \cdashline{3-3}  \cdashline{4-7}           % --
&& $\diamond$ Logotype with accessory  & \checkmark & \checkmark & \checkmark & \checkmark     \\  % Figurative/Shape 
\hline
& \textbf{Emblem}                     & \textbf{Logo-symbol}                & \checkmark & \checkmark & \checkmark & \checkmark     \\   \cdashline{2-7}  % Figurative/Shape
& --                                  & \textbf{Logotype with symbol}       & \checkmark & \checkmark & \checkmark & \checkmark     \\   \cdashline{2-7}  % Figurative/Shape
\textbf{Symbolic}       & \textbf{Only symbol:}	& \textbf{Only symbol:}               &                                                       \\  %\cdashline{2-4} %
\textbf{identifier}     & $\diamond$ Pictorial mark           & $\diamond$ Iconic symbol            & \checkmark & \checkmark & --         & --             \\ \cdashline{2-7}  % Figurative
& $\diamond$ Abstract/symbolic mark   & $\diamond$ Abstract symbol	& --         & \checkmark & \checkmark & --             \\ \cdashline{2-7}  % Shape
& $\diamond$ Letterform               &  $\diamond$ Alphabetic symbol       & --         & \checkmark & \checkmark & --             \\  % Shape
\hline
\end{tabular}
\end{adjustbox}
\end{table}

In this work, the modified Vienna classification is used to perform MLC and similarity search. We will specifically use the dataset provided by EUIPO (described in Section \ref{sec:dataset}), which, in addition to Vienna, uses the alternative Nice classification\footnote{\url{https://euipo.europa.eu/ohimportal/en/nice-classification}} to label goods and services. This categorization organizes the sector into 45 subcategories. The labels used for goods include chemicals, medicines, metals, materials, machines, tools, vehicles, instruments, etc., while the labels used for services include advertising, insurance, telecommunications, transport, and education.

% ---------------------------------------------------------------
\section{Method}
\label{sec:method}

\begin{figure}[ht]
\centering
\includegraphics[width=1\textwidth]{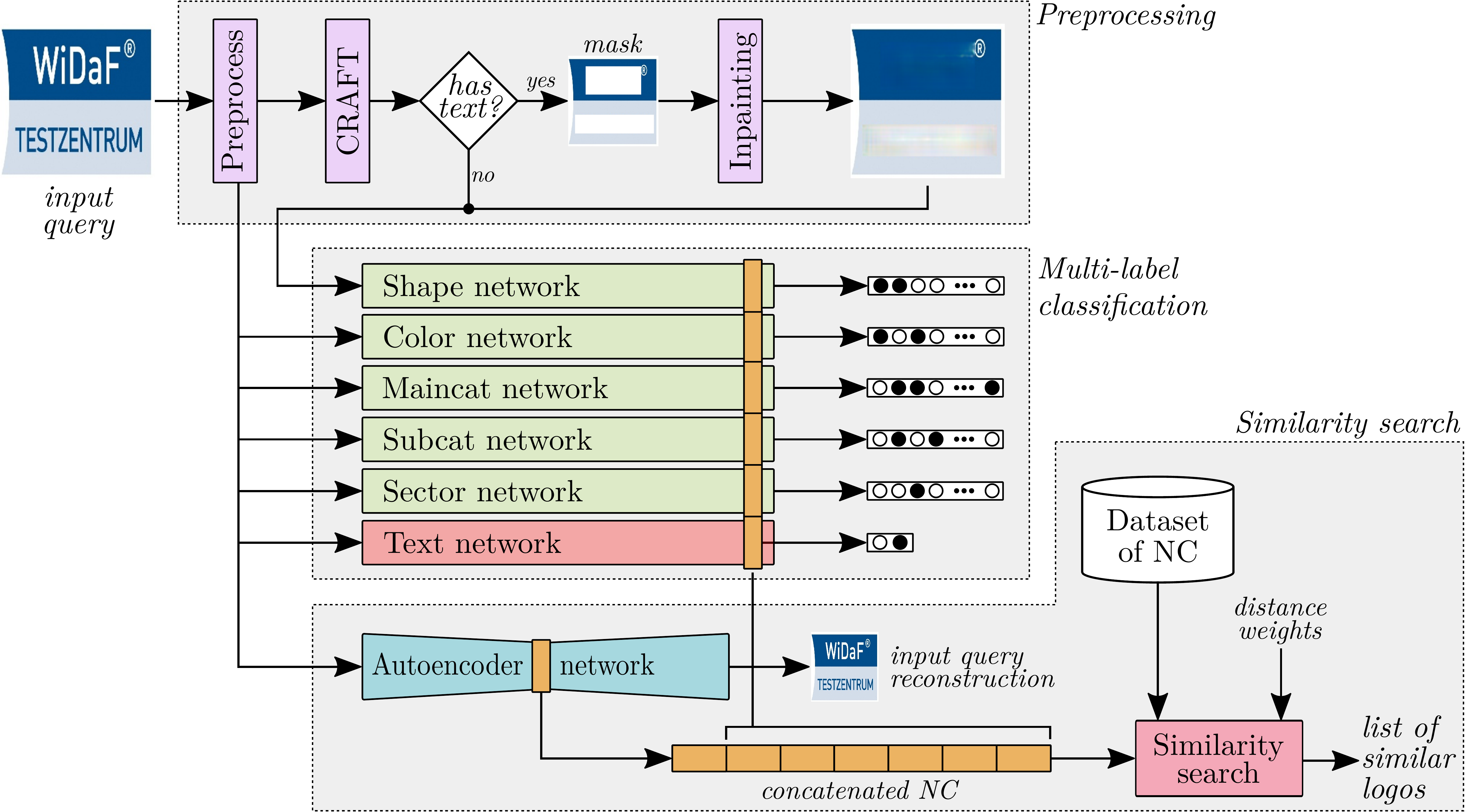}
\caption{Scheme of the proposed method.}
\label{fig:scheme}
\end{figure}

Figure \ref{fig:scheme} shows the scheme of the proposed approach, which is divided into three main steps: a preprocessing of the input images, a multi-label classification, and a similarity search step based on the features learned in the previous stage. Detailed explanations of each of these steps are provided in the following sections.

\subsection{Data preprocessing}

Data preprocessing is performed to prepare the image for the next steps. The logo is first cropped to eliminate the borders containing a uniform background. Logo images used for trademark registration or similarity search (i.e., as long as it is not for the task of searching logos in the wild\footnote{This term refers to the task of finding the position of a logo from a generic image that may contain many other elements.}) generally tend to have a uniform background. Therefore, the images are cropped by eliminating color-uniform borders so that the logo will occupy all the available space in the image. This makes it possible to homogenize their size and facilitate the comparison process.

The second preprocessing step consists of detecting whether the input logo contains text and, if so, generating a new version of it without text. Many image brands include text. However, this information may be irrelevant or even confuse the detection of some logo characteristics. During experimentation, it was observed that shape classification improved notably when the text was eliminated. This was not the case with the other characteristics, such as color or figurative elements. This process, therefore, was carried out only for the shapes, using the full logo for the rest of the characteristics, as shown in Figure~\ref{fig:scheme}.

\begin{figure}[ht]
\centering
\includegraphics[width=0.85\textwidth]{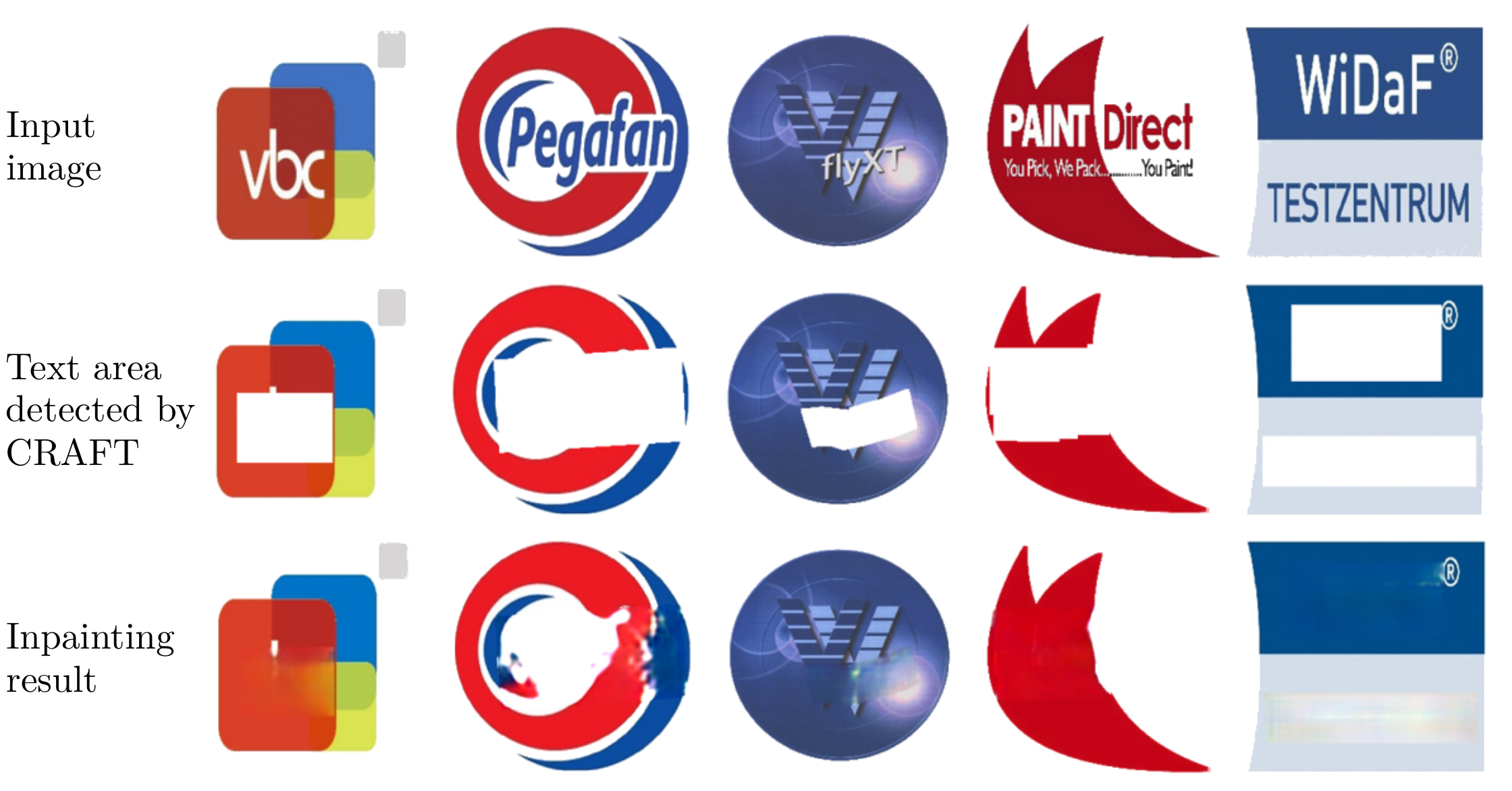}
\caption{Examples of how selected text is removed from the image using CRAFT and how an inpainting neural network fills gaps.}
\label{fig:inpainting_sample}
\end{figure}

To remove the text, the image is first processed using the CRAFT text detector~\citep{baek2019character}, which efficiently identifies the text area of an image by exploring each region and the affinity between text characters. As a result, if any text is found, a mask is obtained. Together with the original image, this mask is processed by an inpainting network \citep{wang2018image} to fill the detected gaps with a background color. For optimization, when the detected mask is surrounded by white pixels ---which is quite common in these kinds of images--- the gap is filled directly to white. Figure \ref{fig:inpainting_sample} shows some examples of the steps followed in this process.

\subsection{Multi-label classification}

In the second step of the proposed method, a set of neural networks is used for the classification of different characteristics of the input image. Specifically, each of these networks specializes in the multi-label classification of one of the proposed label groups (see Section \ref{sec:vienna}), such as shape, color, text, category, subcategory, and sector. 

Figure \ref{fig:scheme} shows a diagram of the integration of these networks into the proposed methodology. The input used is the preprocessed result of the previous stage (in the case of the network specialized in shape, the version of the image without text is employed). The fact that they are independent allows the networks to be run in parallel, signifying that the algorithm's performance is not affected.

\begin{figure}[ht]
\centering
\includegraphics[width=1\textwidth]{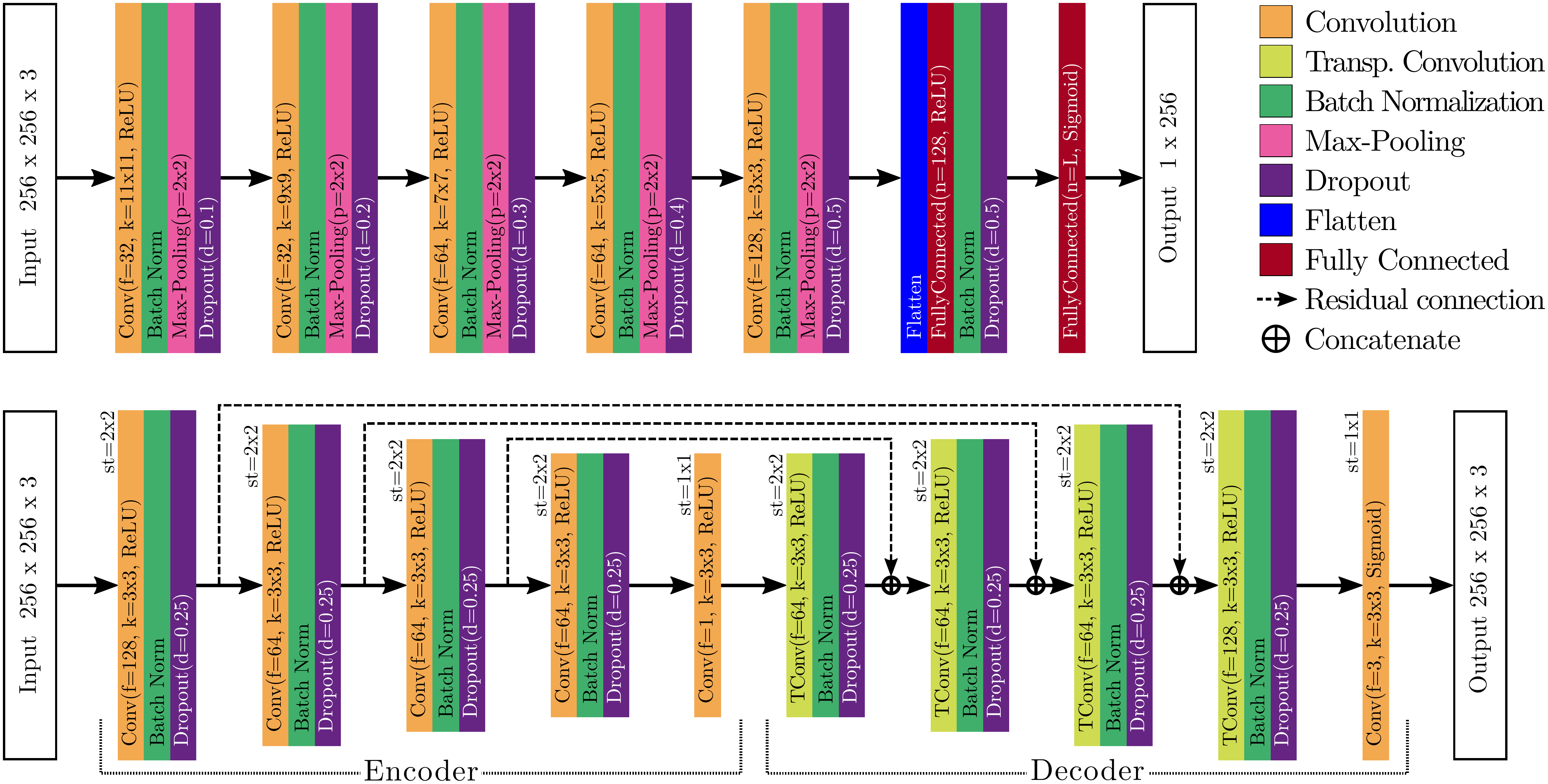}
\caption{Schemes of the specialized CNNs (top, used in the MLC stage) and the Auto-Encoder (bottom, used for the similarity search). In this figure, the layer type is labeled with colors according to the side legend. Each layer configuration is shown in the scheme, including the activation function, the number of filters ($f$) and kernel size ($k$) for convolutions and transposed convolutions, the pool size ($p$) for max-pooling, the ratio $d$ used for dropout, the stride $st$ applied to each layer of the auto-encoder, and the number of neurons $n$ used for the fully-connected layers. 
}
\label{fig:topologies}
\end{figure}

As discussed in the introduction, the current methods that obtain the best results as regards processing logos, or images in general, are those based on CNN \cite{lecun2010, lecun2015deep}, and it is for this reason that we also use this type of architecture, but adapting it to an MLC configuration. The specific definition of the networks used is shown in Figure \ref{fig:topologies} (upper diagram). The proposed architecture consists of five layers alternately arranged into convolutions, batch normalization \citep{BatchNormalization}, max-pooling and dropout \cite{Srivastava2014}, plus two final fully-connected layers, also with dropout. Batch normalization and dropout \cite{Srivastava2014} were included to reduce overfitting, help perform faster training, and improve accuracy. 

ReLU~\cite{Glorot2011relu} was used as the activation function for all layers except the output, which has a sigmoid activation function. The sigmoid function models the probability of each class as a Bernoulli distribution, in which each class is independent of the others, unlike that which occurs with Softmax. Therefore, the output is a multi-label classification for each of the $L$ characteristics considered, which depends on the number of classes of each particular network.

In the case of the network specialized in text detection, only one output is necessary since it detects only the presence of text in the image. Unlike CRAFT, which searches for individual characters, this network seeks global features that allow this binary classification to be carried out. 
As we will see in the experimentation section, this difference has the advantage of allowing a generic comparison of the presence of text in the image (and not of the specific text that appears in it).

%--------------------------------------------
\subsection{Similarity search}

The last step of the method takes advantage of the intermediate representation learned by the networks described in the previous section to perform the logo-similarity search. In this respect, it is possible to use the CNN as a feature extractor to obtain a suitable mid-level vector representation (also called a Neural Code or NC~\cite{Babenko:2014}) that is later used as the input for a search algorithm such as $k$NN~\cite{Gallego2020}. This is done by feeding the trained networks with the raw data and extracting the NC from one of the last layers of the network~\cite{Huang06,Razavian2014}, in our case, from the penultimate layer. 

In addition to the six specialized networks used in the previous step, an auto-encoder architecture is added to capture generic characteristics that define logos. These networks were proposed decades ago by Hinton et al.~\cite{hinton1994autoencoders} and have since been actively researched~\cite{Baldi2012autoencoders}. They consist of feed-forward neural networks trained to reconstruct their input. They are usually divided into two stages: the first part (denominated as the \textit{encoder}) receives the input and creates a meaningful intermediate or latent representation of it, and the second part (the \textit{decoder}) takes this intermediate representation and attempts to reconstruct the input.

Figure \ref{fig:topologies} (bottom) depicts the topology of the auto-encoder used. The encoder consists of four convolutional layers combined with batch normalization and dropout. Down-sampling is performed by convolutions using strides rather than resorting to pooling layers. Four mirror layers are then followed to reconstruct the image to the same input size. Up-sampling is achieved through transposed convolution layers, which perform the inverse operation to a convolution to increase rather than decrease the resolution of the output. Residual connections were also added from each encoding layer to its analogous decoding layer, thus facilitating convergence and improving the results.

The size of the neural codes (NC) obtained from these networks is 128 for the CNNs and 256 for the auto-encoder. In preliminary experiments, it was observed that the accuracy decreased with smaller sizes and that larger sizes did not lead to any improvement. As shown in Figure \ref{fig:scheme}, these NC are combined into a single feature vector, which is then used to perform the similarity search. An $\ell_2$ normalization~\cite{ZhengZWWT16} is applied for the regularization of this vector since this technique usually improves the results~\citep{Gallego2018knn}. 

During the training stage, the NCs from the training set are extracted and stored following the process described above. Then, in the inference phase, the NC representation of the query is obtained and compared with the stored NCs. In this process, in addition to using $k$NN~\cite{DudaHS01}, the result obtained was also compared with the following two multi-label similarity search methods:

\begin{itemize}
    \item \textbf{Binary Relevance $k$NN (BR$k$NN)} \cite{EleftheriosSpyromitros2008}: This is a multi-label classifier based on the $k$NN method and the Binary Relevance (BR) problem transformation. It learns one binary classifier for each different label by checking whether samples are labeled with the label under consideration, thus following a one-against-all strategy. 
    
    \item \textbf{LabelPowerset} \citep{MLCscene}: This also follows a problem transformation approach in which the multi-label set is transformed into a multi-class set. Then, a classifier (Random Forest, in this case) is trained on all the unique label combinations found in the training data.
\end{itemize}

Moreover, we use a weighted distance to search for the nearest neighbors. The advantage of this distance is that it allows the user to adjust the search criteria by modifying the weight assigned to each characteristic (e.g., the user can tune the method to give a higher weight to the shape or color when seeking the most similar logos). The following weighted dissimilarity metric $d_w$ was used to calculate the distance between two vectors $A$ and $B$:

\begin{equation}
d_w(A,B) = \frac{\underset{c \in {\cal C}}\sum w^c d(A^c,B^c)}{\underset{c \in {\cal C}}\sum w^c}
\label{eq:wdist}
\end{equation}

\noindent
where ${\cal C}$ is the set of all possible characteristics (i.e., color, shape, etc.), $A^c$ and $B^c$ represent the subset of features corresponding to the characteristic $c$, $w^c$ is the weight assigned to that characteristic, $\forall c \in {\cal C}: w^c \in [0,1]$, $\sum_{c \in {\cal C}}w^c=1$, and $d: \mathcal{X}\times\mathcal{X}\rightarrow\mathbb{R}^{+}_{0}$ is the dissimilarity metric used to compare the two vectors. We have employed the Euclidean distance since, as described above, the NC vectors obtained are numerical feature representations.

\subsection{Training process}

The training of the networks was made using standard back-propagation, Stochastic Gradient Descent (SGD)~\cite{Bottou2010}, and considering the adaptive learning rate method proposed in \cite{Zeiler12}. The \emph{binary crossentropy} loss function was used to calculate the error between the CNN output and the expected result. The training lasted a maximum of 100 epochs with a mini-batch size of 32 samples and \emph{early stopping} when the loss did not decrease during 15 epochs. 

A model pretrained during 25,000 iterations with more than 10,000 images was used for the CRAFT~\cite{baek2019character} network. The inpainting network \cite{wang2018image} was initialized with a model pretrained with ImageNet and fine-tuned with our dataset during 30,000 iterations.

%-------------------------------------------------------------------------------------------------
\section{Experimental setup} 
\label{sec:setup}

\subsection{Dataset}
\label{sec:dataset}

The experimentation was carried out using the European Union Trademark (EUTM) dataset provided by EUIPO\footnote{\url{https://euipo.europa.eu/ohimportal/en/open-data}}. This dataset is labeled using the Vienna classification as depicted in Section \ref{sec:topologies}. However, since the available labeling is not exhaustive, a filtering process was performed to select only those logos whose semantics, color, and shape were labeled. We, therefore, eventually chose a subset of 76,000 logos corresponding to the 2010-2018 period.

\begin{figure}[ht]
\centering
\includegraphics[width=1\textwidth]{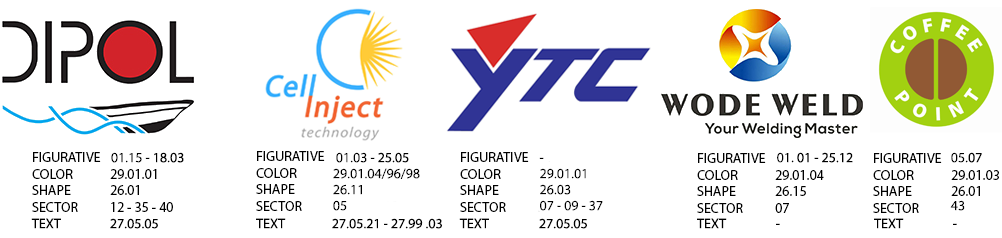}
\caption{Some examples of trademarks in the EUTM dataset. Note that some of them have only partial labeling of some characteristics, such as color and shape, and that the text is not labeled although it is present, as it is not considered to be a characteristic element of the design.}
\label{fig:data_sample}
\end{figure}

It is important to state that even if this filtering is performed, the collected labeling is still not complete. This is owing to the subjectivity of some labels and the fact that operators usually indicate only the most representative characteristics of logos, i.e., those that are distinctive of that brand. For example, in the first image in Figure \ref{fig:data_sample}, it will be noted that only the color red was labeled, although it also contains black and blue. The same happens with the third, fourth, and fifth logos, in which only one color is labeled although they contain more.
%In the third logo, green and black were labeled, while in the fourth and fifth, which have three colors, only one of them was detailed. 
%In the case of the shape labeling, only circles were annotated in the first, third, and fifth images, although they also contain other shapes.
In the case of the shape labeling, only circles were annotated in the first and fifth images, lines for the second logo, and triangles for the third, although they also contain other shapes.

In the case of text labeling, only 30 \% of the images had this information. Again, in this dataset, the presence of text is labeled only when it is a distinctive element. 
%For example, Figure \ref{fig:data_sample} shows that the text is labeled only in the first image when three other logos contain text. 
For example, Figure~\ref{fig:data_sample} shows that the text is only labeled in the first three images when all the logos contain text. 
For this reason, all the images were processed and reviewed using the CRAFT text detector to complete this labeling, thus obtaining much more complete ground truth for this feature.

The input images were scaled to a spatial resolution of 256$\times$256 pixels, and their values were normalized into the range $[0, 1]$ to feed the networks. Of the 76,000 logo images, 80\% were selected for training, and the remaining samples were employed for testing.

\subsection{Metrics}

In multi-label learning, each sample may have more than one ground-truth label. To assess this problem quantitatively, better ranks are assigned as the method correctly predicts more ground truth labels. In this work, we considered the following two multi-label metrics~\cite{Tsoumakas10miningmulti-label}. 

\subsubsection*{Label Ranking Average Precision (LRAP)} 
This is a label ranking (LR) metric that is linked to the average precision score but based on the notion of label ranking rather than precision and recall. LRAP averages over the samples the answer to the following question: for each ground truth label, what fraction of higher-ranked labels were true labels? This performance measure will be higher if the method can give a better rank to the labels associated with each sample. The score is always strictly greater than 0, with 1 being the best score. 

Formally, given a binary indicator matrix of the ground truth labels, $y \in \{0,1\}^{N\times{L}}$, where $N$ and $L$ are the amount of samples and labels, respectively, and the score associated with each label $\hat{f} \in \mathbb{R}^{N\times{L}}$, the LRAP is defined as: 

\begin{equation}
\textrm{LRAP}(y,\hat f) = \frac{1}{N} \sum_{i=0}^{N-1} \frac{1}{{\|y_i\|}_0} \sum_{j:y_{ij}=1}\frac{| \mathcal{L}_{ij} |}{\mathrm{rank}_{ij}}
\end{equation}

\noindent 
where $\mathcal{L}_{ij} = \left\{k: y_{ik} = 1, \hat{f}_{ik} \geq \hat{f}_{ij} \right\}$, $\mathrm{rank}_{ij} = \left|\left\{k: \hat{f}_{ik} \geq \hat{f}_{ij}\right\}\right|$, $|\cdot |$ calculates the cardinality (number of elements) of the set, and ${\|\cdot\|}_0$ is the $\ell_0$-norm that computes the number of nonzero elements in a vector. If there is exactly one relevant label per sample, LRAP is equivalent to the Mean Reciprocal Rank (MRR).

\subsubsection*{Label Ranking Loss (LRL)} 
This LR metric computes the ranking, which averages the number of label pairs that are incorrectly ordered in the samples (true labels with a lower score than false labels), weighted by the inverse of the number of ordered pairs of false and true labels. The best performance is achieved with an LRL of zero. This metric is formally defined as: 

\begin{equation}
\textrm{LRL}(y,\hat f) = \frac{1}{N} \sum_{i=0}^{N-1} \frac{1}{\|y_i\|_{0}(L - \|y_i\|_{0})} \left|\left\{(k, l): \hat{f}_{ik} \leq \hat{f}_{il}, y_{ik} = 1, y_{il} = 0 \right\}\right|
\end{equation}

% ---------------------------------------------------------------------------
\section{Evaluation}
\label{sec:evaluation}

The proposed methodology is evaluated at different levels, starting with the MLC phase and continuing with the similarity search, also comparing it with other state-of-the-art approaches. In addition to quantitative results, a qualitative evaluation is carried out by analyzing the response of each stage of the method and comparing the results with the classification made by experts and graphic design students.

\subsection{Multi-label classification}

At this stage, the method returns a multi-label classification for each characteristic considered, i.e., color, shape, main category, sub-category, and sector. Table \ref{tab:mlc_results} shows the results obtained for each of these characteristics in terms of LRAP and LRL. There is a consensus regarding the best and worst results, with color, the main category, and the sub-category being the best-detected characteristics. The worst-ranked feature is the sector. This is because no specific pattern, characteristic, or type of design is detected with which to determine it since the type of design applied to each sector is  subjective.

Intermediate precision was attained for shape classification, mainly owing to the labeling noise and the ambiguity of the possible classes. In these results, there is also an improvement produced by the proposed preprocessing to eliminate the text from the image (``Shape+'' row) compared to using the original version of the logo that includes the text (``Shape'' row).

The results for the text classification network are not included in this table since this is not a multi-label classifier (it discriminates only whether or not the image contains text). For this reason, the accuracy metric was chosen, obtaining 96.06\% for this task.

\begin{table}[ht]
\setlength{\tabcolsep}{9pt}
\caption{Results obtained with the proposed method for the multi-label classification stage. Two cases are shown for the Shape network: ``Shape+'', which includes the preprocessing to remove the text, and ``Shape'', which does not. 
}
\label{tab:mlc_results}
\centering
\begin{footnotesize}
\begin{tabular}{lcc}
\hline
\textbf{Model}          & \textbf{LRAP}     & \textbf{LRL}          \\ 
\hline
\textbf{Color}		& 0.8642		& 0.0561                \\
\textbf{Sub-category}    & 0.7376		& 0.0561                \\
\textbf{Main category}  & 0.7979		& 0.0635                \\
\textbf{Shape+}	& 0.7699		& 0.1169                \\
\textbf{Shape}		& 0.6899		& 0.1534                \\
\textbf{Sector}		& 0.8890		& 0.2220                \\
%\textbf{Subsector}	& 0.3165		& 0.2642                \\

\hline
\end{tabular}
\end{footnotesize}
\end{table}

Table \ref{tab:mlc_examples} depicts some examples of the results obtained for multi-label classification, including only the predictions made with a confidence greater than 2\%. When the prediction is compared with the ground truth (GT), the method succeeds in all cases, with a fairly high confidence percentage. The only error was in the main category of the second logo since the class ``\textit{plants}'' was selected as the first option. However, this error is understandable when the examples labeled with this class are analyzed since they usually are green and define shapes with curves. For the shape labeling, it can be seen that in some examples, such as the first, third and fourth, other classes that were not labeled are proposed but that, nevertheless, describe characteristics present in the logos.

\begin{table}[ht]
\renewcommand{\arraystretch}{1.2}
\setlength{\tabcolsep}{6pt}
\caption{Examples of multi-label classification of the EUTM dataset, including the ground truth (GT) and the prediction made by the main category, shape, color, and text networks when the confidence percentage of the prediction exceeds 2\%. Two examples include text, and two do not.}
\label{tab:mlc_examples}
\centering
\begin{adjustbox}{width=\columnwidth, keepaspectratio}
\begin{tabular}{llllll}
\hline
&&
\includegraphics[width=.17\textwidth]{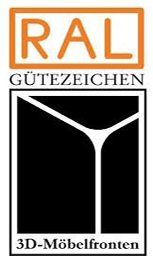} &
\includegraphics[width=.24\textwidth]{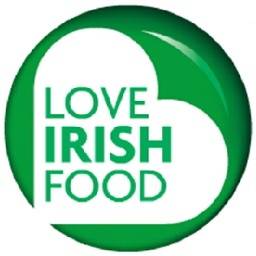} &
\includegraphics[width=.22\textwidth]{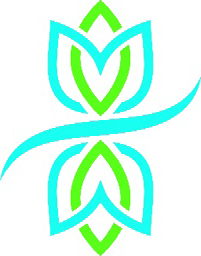} &
\includegraphics[width=.24\textwidth]{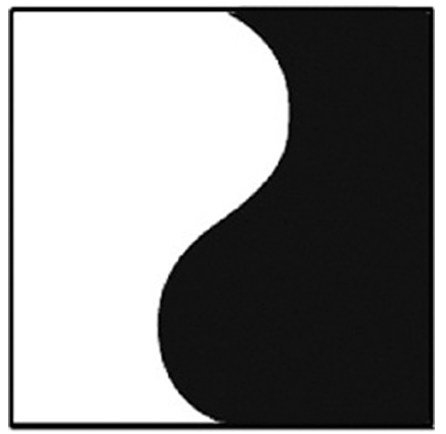} \\
\hline
\multirow{6}{*}{\rotatebox[origin=l]{90}{\textbf{GT}}}  
& \textbf{Main-category:} 
& Ornamental motifs
& Human beings
& Plants
& \begin{tabular}[l]{@{}l@{}}Ornamental motifs\\Games, toys\end{tabular} \\ \cdashline{2-6}
& \textbf{Shape:} 
& \begin{tabular}[l]{@{}l@{}}Quadrilaterals\\Lines, bands\end{tabular} 
& Circles, ellipses
& Lines, bands
& Quadrilaterals \\ \cdashline{2-6}
& \textbf{Color:}
& Black; Orange
& Green
& Blue
& Black; White \\ \cdashline{2-6}
& \textbf{Text:}
& Yes
& Yes
& No
& No
\\ 
\hline
\multirow{10}{*}{\rotatebox[origin=l]{90}{\textbf{Prediction}}} 
& \textbf{Main-category:}
& 100\% ornamental motifs
& \begin{tabular}[l]{@{}l@{}}
47.10\% plants \\
17.70\%	human beings \\
4\% arms, ammunition
\end{tabular} 
& \begin{tabular}[l]{@{}l@{}}
46.47\% plants \\
31\% heraldry, coins \\
9.66\% celestial bodies 
\end{tabular} 
& 100\% ornamental motifs \\ \cdashline{2-6}
& \textbf{Shape:}
& \begin{tabular}[l]{@{}l@{}}
94.85\% quadrilaterals \\
6.07\% lines, bands \\
4.08\% other polygons
\end{tabular}  
& 99.81\% circles, ellipses
& \begin{tabular}[l]{@{}l@{}}
63.62\% circles, ellipses \\
61.87\% lines, bands \\
10.44\% quadrilaterals
\end{tabular} 
& \begin{tabular}[l]{@{}l@{}}
99.96\% quadrilaterals \\
10.06\% circles, ellipses
\end{tabular} 
\\ \cdashline{2-6}
& \textbf{Color:}
& \begin{tabular}[l]{@{}l@{}}
48.44\% black \\
94.41\% orange \\
58.45\% white
\end{tabular}
& 99.22\% green
& 100\% blue
& \begin{tabular}[l]{@{}l@{}}
87.54\% black \\
78.36\% white
\end{tabular} 
\\ \cdashline{2-6}
& \textbf{Text:}
& Yes
& Yes
& No
& No \\
\hline
\end{tabular}
\end{adjustbox}
\end{table}

\subsection{Similarity search}

In this section, we evaluate the similarity search results using the NCs learned by the neural networks in the MLC stage together with the NCs of the auto-encoder. In this case, the results are reported by considering only the LRAP metric since, as stated in the previous section, the tendency of both metrics is similar.

To establish the value of $k$ used by the $k$NN and BR$k$NN while simultaneously analyzing the labeling noise, we shall now evaluate the result obtained when performing the similarity search for the single-label case. For this, when processing each class, only the samples with a single label for that characteristic were considered. Table \ref{tab:knn_results} depicts the results of this experiment (in terms of LRAP) when considering the $k$NN method and values of $k$ in the range $[1, 11]$. As will be noted, the best results are obtained with high $k$ values, between 7 and 11. The intermediate value of $k=9$ was eventually chosen for the remaining experiments. These results demonstrate that the labels provided contain noise since the method improves by considering more neighbors in the inference stage. 

\begin{table}[ht]
\setlength{\tabcolsep}{4pt}
\caption{Similarity search results (in terms of LRAP) obtained with the $k$NN classifier for the single-label search task and different $k$ values. The best results are highlighted in bold type.}
\label{tab:knn_results}
\centering
\begin{footnotesize}
\begin{tabular}{lcccccc}
\hline
\textbf{Model}          & \multicolumn{6}{c}{\textbf{CNN + $k$NN}}  \\ \cline{2-7}
& \textbf{k=1}  & \textbf{k=3}  & \textbf{k=5}  & \textbf{k=7}  & \textbf{k=9}  & \textbf{k=11} \\ 
\hline
%\textbf{Text}		& 0,9862        & 0,9866        & \textbf{0,9868} & 0,9866      & 0,9866        & 0,9867        \\
\textbf{Color}		& 0.8322        & 0.8366        & 0.8369        & \textbf{0.8394} & 0.8378        & 0.8378        \\
\textbf{Main Category}  & 0.7673        & 0.7842        & 0.7875        & 0.7885        & \textbf{0.7886} & 0.7880         \\
\textbf{Subcategory}    & 0.7409        & 0.7611        & 0.7660        & 0.7682        & 0.7695        & \textbf{0.7716}   \\
\textbf{Sector}		& 0.8020        & 0.8027        & 0.8054        & 0.8060        & 0.8065        & \textbf{0.8067}         \\
%\textbf{Subsector}	& 0.0883        & 0.0553        & 0.0474        & 0.0454        & 0.0423        & 0.0397        \\
\textbf{Shape}		& 0.5489        & 0.5513        & 0.5503        & 0.5542        & 0.5544        & \textbf{0.5552}   \\
\textbf{Shape+}	& 0.6583        & 0.6717        & 0.6707        & 0.6689        & 0.6712        & \textbf{0.6728}   \\
\hline
\end{tabular}
\end{footnotesize}
\end{table}

Since LabelPowerset is based on Random Forests, we also carried out a similar experiment by evaluating the number of trees considered in the range $t \in [100,500]$, eventually obtaining the best result with $t=100$. These parameter settings were used to compare the three multi-label similarity search algorithms: $k$NN and BR$k$NN with $k=9$, and LabelPowerset with $t=100$. Table \ref{tab:sim_results} shows the results of this experiment using the LRAP metric. 
%The best result per method is highlighted in bold type. 
As can be seen, a better result is obtained for almost all the characteristics when using LabelPowerset. The only exception is the sector, which, as previously argued, is a very subjective characteristic and may contain a higher level of noisy labels.

\begin{table}[ht]
\setlength{\tabcolsep}{9pt}
\caption{Results obtained for the different characteristics with the three multi-label classifiers using the LRAP metric. The best results are shown in bold type.}
\label{tab:sim_results}
\centering
\begin{footnotesize}
\begin{tabular}{lccc}
\hline
\textbf{Network}        &\textbf{$k$NN}    &\textbf{BR$k$NN}    &\textbf{LabelPowerset}          \\ \hline
%\textbf{Text}		& 0,9866                & 0,9865		& 0,9870                \\
\textbf{Color}		& 0.7042                & 0.7042		& \textbf{0.7070}       \\
\textbf{Main Category}	& 0.7015                & 0.7015		& \textbf{0.7396}       \\
\textbf{Subcategory}    & 0.6589                & 0.6589   	& \textbf{0.6850}       \\
\textbf{Sector}		& \textbf{0.8434}       & \textbf{0.8434}  & 0.8001                \\
\textbf{Shape}		& 0.5333                & 0.5333		& \textbf{0.5594}       \\
\textbf{Shape+}	& 0.6242                & 0.6242   	& \textbf{0.6579}       \\
\hline
\end{tabular}
\end{footnotesize}
\end{table}

\begin{figure}[ht]
\centering
\includegraphics[width=.5\linewidth] {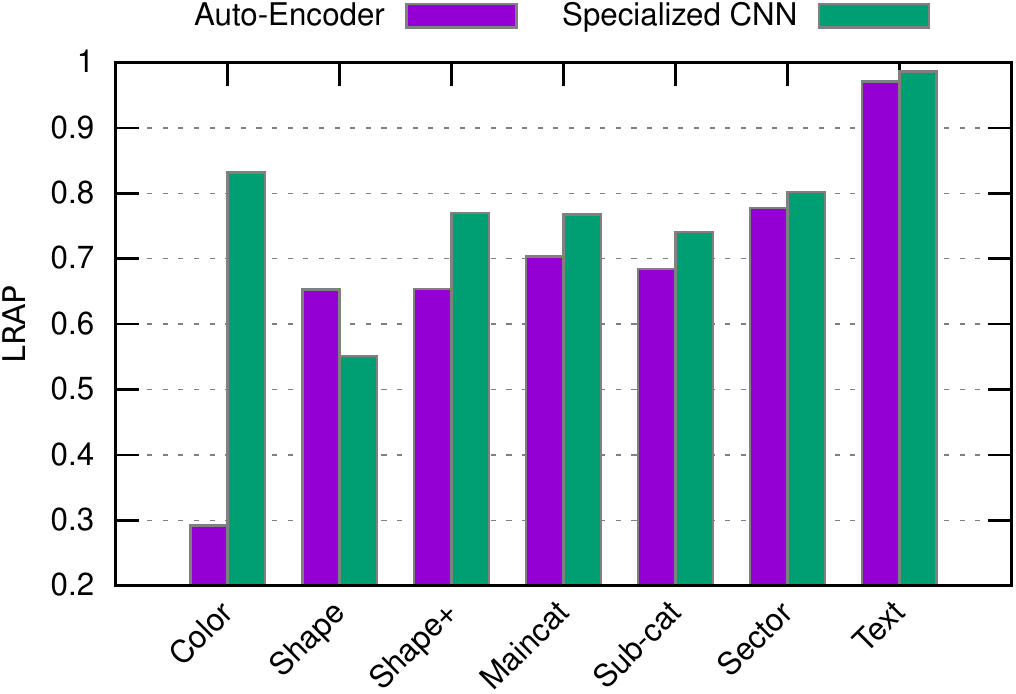}
\caption{Similarity search results obtained for each of the characteristics considered when using the NCs learned by the auto-encoder. The results obtained by the specialized networks for these characteristics are included as a reference.}
\label{fig:knn_ae}
\end{figure}

In the case of the auto-encoder, it is necessary to consider that it is trained in an unsupervised manner for the reconstruction of the input. The labeling of characteristics is, therefore, not used during training. 
For this reason, to assess its performance, its result for the characteristics considered is compared with that obtained by the specialized networks. 
%For this reason, we evaluate its performance for all the characteristics considered, comparing the result with that obtained by the specialized networks. 
Figure \ref{fig:knn_ae} shows this analysis. Good results are yielded for almost all characteristics except for color. This indicates that the auto-encoder learns a generic representation of combined features, which primarily considers shape versus color.

It is also evident that the auto-encoder works even better than the shape network when the text is not eliminated, but this is not the case when the proposed preprocessing (Shape+) is applied. The auto-encoder is not the best for any particular feature (except for shape without preprocessing). This method is, therefore, beneficial for searching for similarity generically, considering appearance without looking at any specific characteristic.

\subsubsection{Qualitative results}

\begin{figure}[ht]
\centering
		\includegraphics[width=1\linewidth]{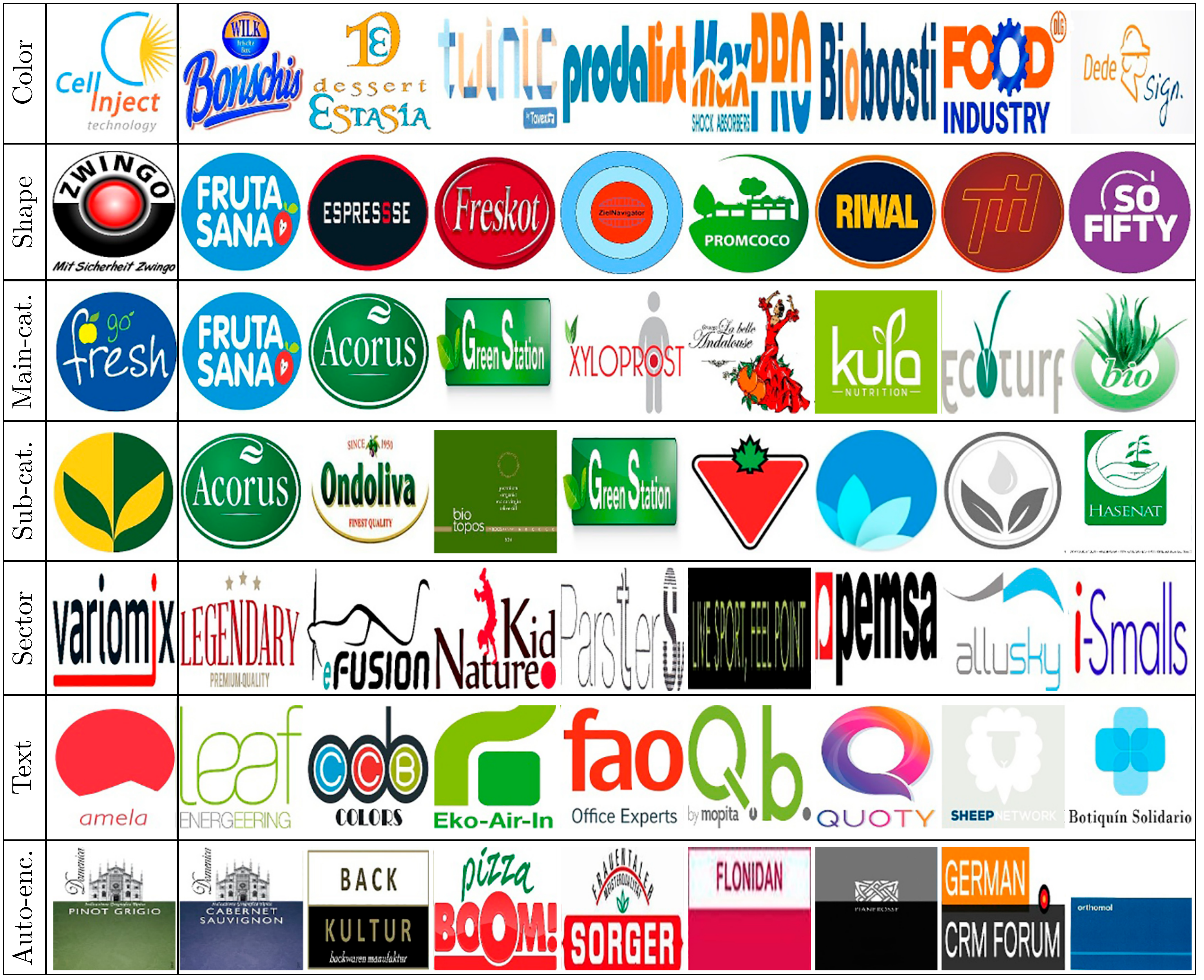}
		\caption{Example of the 8-nearest neighbors obtained by using each of the specialized networks separately, that is, assigning 100\% of the search weight to a single feature. The first logo is the query.}
	\label{fig:examples_sim}
\end{figure}

In this section, we qualitatively analyze the results obtained after the similarity search. Figure \ref{fig:examples_sim} includes a series of examples of the logos found when using each specialized network separately, assigning 100\% of the search weight to a single characteristic. In this figure, the first logo in the row is the query, and the others are the 8-nearest neighbors retrieved.

In the case of color (first row of the figure), it will be noted that the results retrieved are correctly matched, even when there are multiple colors, independently of other characteristics such as the shape. The second row depicts an example of shape, which is also perfectly detected without, in this case, taking into account color.

The main category and sub-category (3\textsuperscript{rd} and 4\textsuperscript{th} rows) of figurative designs are more difficult to analyze visually since elements can often be represented creatively or abstractly. It is for this reason that ``\textit{Plants}'' has been selected for the main category and ``\textit{Leaves, needles, branches with leaves or needles}'' for the sub-category, as they contain easily recognizable designs. In both cases, it will be noted that similar logos, in which leaves or plants appear, have been retrieved. For the main category, the design appears to be a little more generic, including other elements such as people, while for the sub-category, the designs are more specific, and only logos that include leaves are shown.

The case of the sector (5\textsuperscript{th} row) is even more difficult to analyze visually since the classification into goods and services is quite subjective and does not always depend on visual information. Nevertheless, this example shows a correct search result for a logo used for goods. In the case of the text (penultimate row), in addition to retrieving logos containing text, the model also considers the logo's composition since a similar design appears in all of them (with the text at the bottom). Finally, the auto-encoder (last row) focuses principally on the spatial distribution or the layout of the logo and, in some cases, also considers the colors.

\begin{figure}[ht]
\begin{footnotesize}
\centering
\begin{adjustbox}{width=\columnwidth, keepaspectratio}
\begin{tabular}{|ll|p{0.8\textwidth}|}
\hline
\multirow{2}{*}{\includegraphics[width=.1\textwidth]{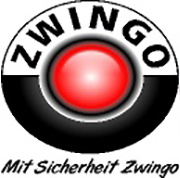}}
& \begin{tabular}[!t]{l}0.7 Color\\0.3 Shape\end{tabular} 
& \begin{tabular}[!t]{l}\includegraphics[width=.8\textwidth]{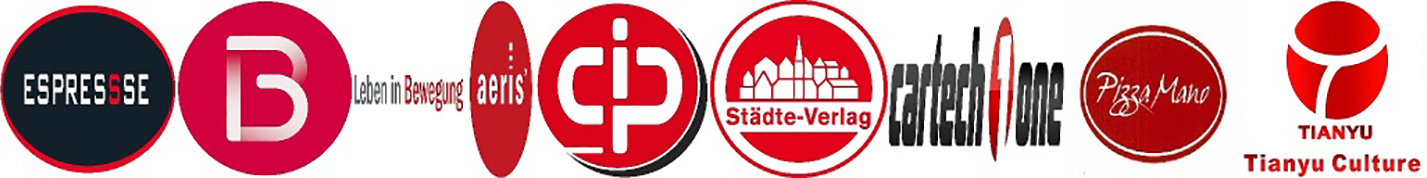} \end{tabular} 
\\
& \begin{tabular}[!t]{l}0.3 Color\\0.7 Shape\end{tabular} 
& \begin{tabular}[!t]{l}\includegraphics[width=.8\textwidth]{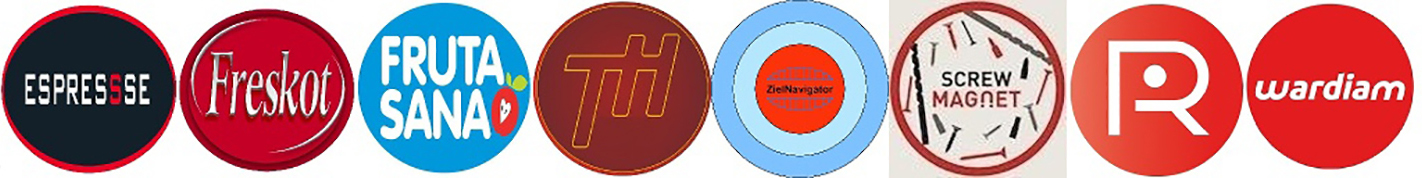} \end{tabular} 
\\
\hline                        
% \multirow{2}{*}{\includegraphics[width=.1\textwidth]{c-1718-query.jpg}}
%     & \begin{tabular}[!t]{l}0.7 Color\\0.3 Shape\end{tabular} 
%     & \begin{tabular}[!t]{l}\includegraphics[width=.8\textwidth]{c07-1718.jpg} \end{tabular} 
%     \\
%     & \begin{tabular}[!t]{l}0.3 Color\\0.7 Shape\end{tabular} 
%     & \begin{tabular}[!t]{l}\includegraphics[width=.8\textwidth]{c03-1718.jpg} \end{tabular} 
%     \\
% \hline
\multirow{2}{*}{\includegraphics[width=.1\textwidth]{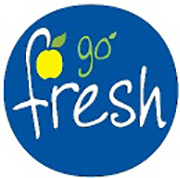}} 
& \begin{tabular}[!t]{l}0.7 Main-category\\0.3 Shape\end{tabular} 
& \begin{tabular}[!t]{l}\includegraphics[width=.8\textwidth]{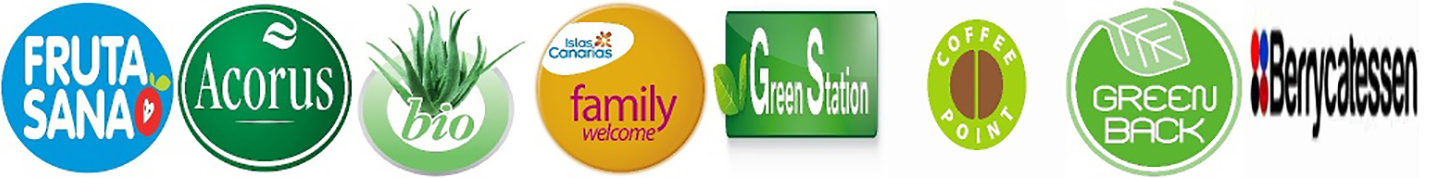}\end{tabular} 
\\
& \begin{tabular}[!t]{l}0.3 Main-category\\0.7 Shape\end{tabular} 
& \begin{tabular}[!t]{l}\includegraphics[width=.8\textwidth]{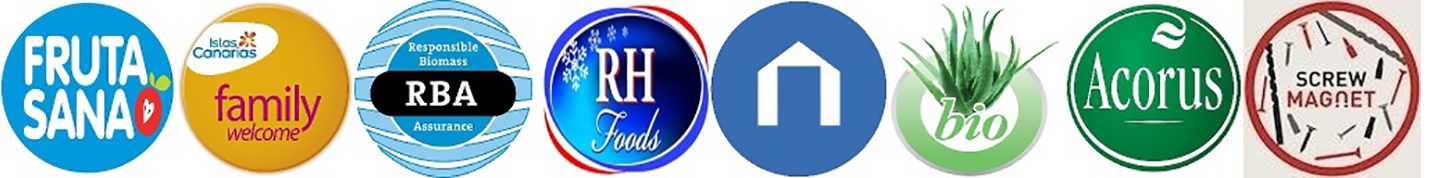}\end{tabular} 
\\
\hline
\end{tabular}
\end{adjustbox}
\end{footnotesize}
	\caption{Results obtained using the weighted distance with two different characteristics. The first column shows the query and the weights applied. The second column includes the 8-nearest neighbors retrieved.
	}
\label{fig:ret_weighted_dist}
\end{figure}

We shall now analyze the effect of combining several characteristics using the proposed weighted distance (see Equation \ref{eq:wdist}) and the capacity it gives users to refine the search. Figure \ref{fig:ret_weighted_dist} shows some examples of the results obtained by applying different weights to combine color, shape, and figurative elements. 

In the first row, the logo used previously in Figure \ref{fig:examples_sim} for the shape characteristic is evaluated, but shape and color are combined in this example. As can be seen, when adding the color, circular logos are again retrieved, but in this case, they have similar colors. When reducing the weight of the color to 30\%, other colors such as blue begin to appear, but red and black are always maintained. These results contrast those previously obtained in Figure \ref{fig:examples_sim}, in which the colors changed completely.

In the second example, the same logo from Figure \ref{fig:examples_sim} (3\textsuperscript{rd} row) is evaluated, but in this case, the figurative elements from the main category are combined with the shape. As will be noted, by giving some weight to shape, the recovered figurative elements keep the same shape, unlike the previous result in which this characteristic was not considered. By assigning 30\% of the weight to the shape, only two logos that do not have a circular shape are obtained, and by giving more weight to the shape (70\%), all the results obtained are ``circles, ellipses''.

\begin{figure}[!ht]
\centering
\includegraphics[width=1\linewidth]{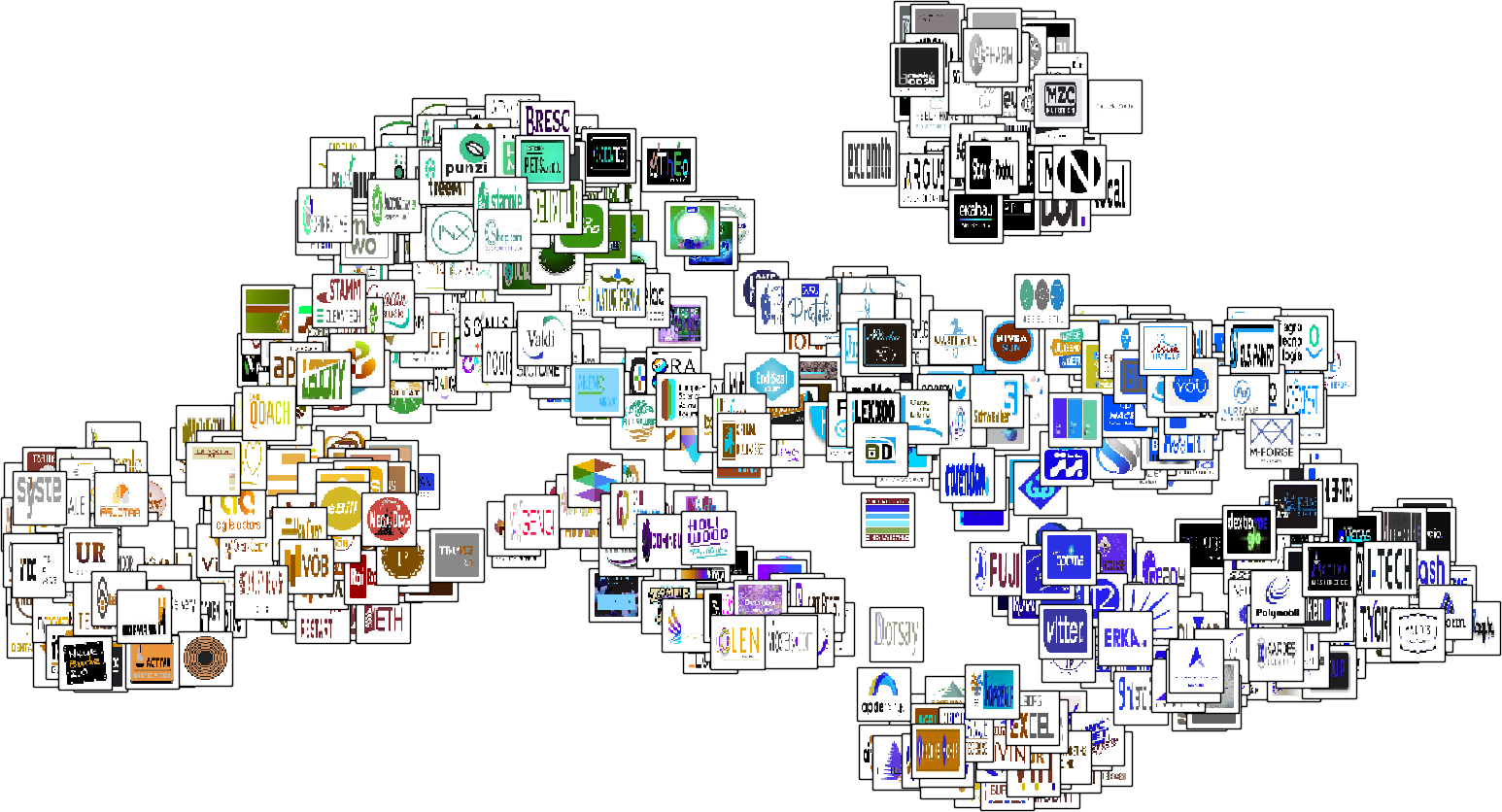}
\\
\includegraphics[width=.45\linewidth]{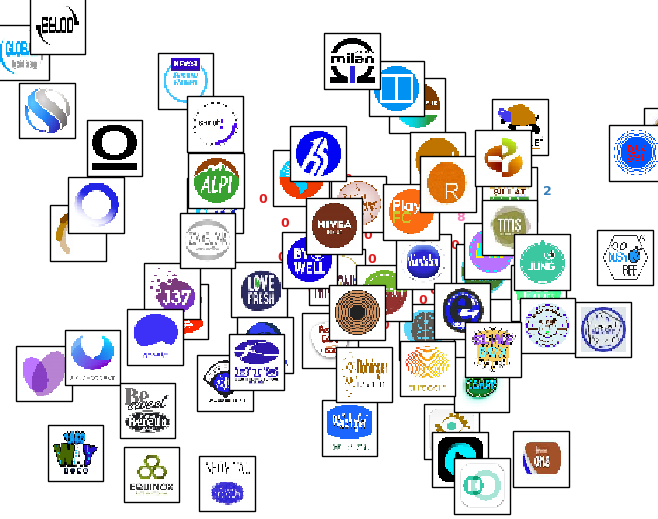}
\hspace{0.9cm}
\includegraphics[width=.45\linewidth]{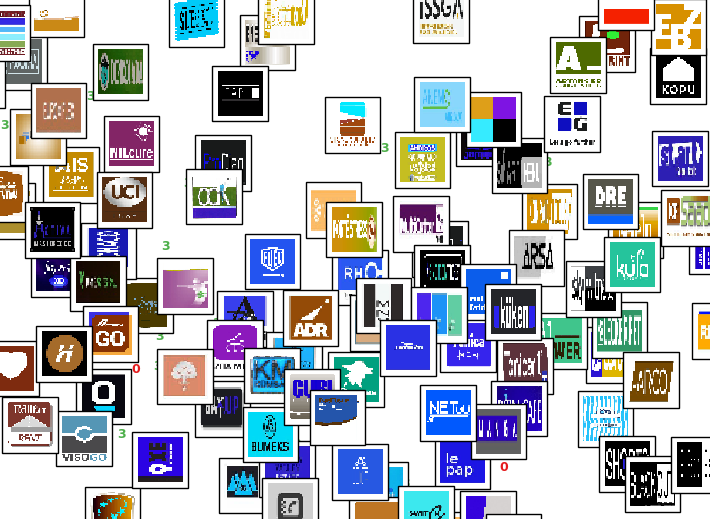}
\caption{Clusters formed by the NCs from the networks of color (top) and shape (bottom) using the t-SNE method. In the case of the shape, two images are included by zooming in on areas in which the circular (left) and quadrilateral (right) shapes are located.}
\label{fig:tsne}
\end{figure}

To analyze the representations learned by the models, a visualization of the grouping formed by the NCs is included for the color and shape characteristics using the t-Distributed Stochastic Neighbor Embedding technique (t-SNE~\cite{Maaten2008}). Figure \ref{fig:tsne} shows that, despite being a multi-label task, the learned NCs tend to group similar characteristics. For example, in the case of the color (top image), gray or silver are grouped in the upper right, blues on the right, yellows, browns, oranges on the left, and greens in the upper left part. Similar shapes are also grouped (see images below, in which two areas of the representation generated are shown zoomed in). In the left-hand image, there are circular shapes, and in the right-hand one, there are quadrilaterals. It should be noted that logos that include text next to these shapes are also grouped separately.

These results show how the networks transform the input images into a new dimensional space (the NCs extracted) in which the logos with similar characteristics are close. This makes it possible to perform a similarity search based on the distance between the representation of the logos in this dimensional space and thus analyze the neighborhood space of a given query to retrieve similar images.

\subsection{Comparison with state of the art}

This section compares the proposed method with other state-of-the-art methods for TIR. Since, and as previously mentioned, there are, to the best of our knowledge, no other MLC approaches for logos that use the Vienna classification, we perform this comparison using METU v2. This dataset is the largest public dataset for TIR and contains 922,926 trademark images belonging to approximately 410,000 companies. Its evaluation set is composed of 417 queries divided into 35 groups of about 10-15 trademarks, in which the logos within the same group are similar.

The evaluation was carried out using the Normalized Average Rank (NAR) metric since it is the measure most commonly employed in reference state-of-the-art works. This metric is calculated by injecting the query set into the main dataset and, for each query logo, the rank obtained for the logos in the same group is calculated as follows:

\begin{equation}
\textrm{NAR} = \frac{1}{N \times N_{rel}} \sum_{i=1}^{N_{rel}}R_i - \frac{N_{rel}(N_{rel} + 1)}{2}
\end{equation}

\noindent where $N_{rel}$ is the number of relevant images for a particular query image (the number of injected images), $N$ is the size of the image set, and $R_i$ is the rank of the $i^{th}$ injected image. The value 0 corresponds to the best performance and 0.5 to a random order. 

Table \ref{tab:cmp_results} shows the result of the comparison carried out. As can be seen, different types of approximations were considered, which were based on both hand-crafted features and neural networks. In the case of those based on hand-crafted features, the use of color histograms \cite{Lei1999}, LBP \cite{Ojala2002MultiresolutionGA}, SIFT \cite{Lowe2004}, SURF \cite{Bay2008}, TRI-SIFT and OR-SIFT \cite{Kalantidis2011} was compared. We also considered two more elaborated proposals: the use of SIFT while excluding the features of the text areas \cite{Perez2018}, and an enhanced version of SIFT \cite{Feng2018} in which reversal invariant features are extracted from edges of segmented blocks which are then aggregated to perform the similarity search.

The use of pre-trained neural network models was also compared. In particular, we evaluated GoogLeNet \cite{Szegedy2015}, AlexNet \cite{Krizhevsky2012} and VGG16 \cite{Simonyan2015}, extracting the NCs from one of its layers (77S1, FC7, and Pool5, respectively). Specific proposals for this dataset were also considered, such as the work of Tursun et al. \cite{Tursun2017}, in which six hand-crafted features are combined with NCs extracted from three different CNN architectures. We also evaluated the proposal of Perez et al. \cite{Perez2018}, which compares three solutions: the results of the VGG19 architecture trained in two ways (one to distinguish visual similarities and the other for conceptual similarities), and the result of merging the features of both. Finally, we included a work based on attention mechanisms \cite{Tursun2020b}, which pays direct attention to critical information, such as figurative elements, and reduces the attention paid to non-informative elements, such as text and background. This process, denominated as ATRHA (Automated Text Removal Hard Attention), is combined with two proposals for the elaboration of the features compared, one based on the Regional Maximum Activations of Convolutions (R-MAC) and the other based on the saliency of Convolutional Activations Maps (CAM) that were detected through the use of soft attention mechanisms (CAMSA) and the aggregation of Maximum Activations of Convolutions (MAC).

As noted in the results shown in Table \ref{tab:cmp_results}, generally, the methods based on neural networks are significantly better than those based on hand-crafted features. There are, however, some exceptions: since the pre-trained networks have not been specifically prepared for this type of data, they do not achieve good results and are even surpassed by a method based on hand-crafted features (``Enhanced SIFT'' \cite{Feng2018}). It is interesting to see how the combination of hand-crafted features with features extracted from a CNN (proposed in \cite{Tursun2017}) achieves a notable improvement. Of the methods based solely on neural networks, the proposal that uses attention mechanisms \cite{Tursun2020b} stands out.

Concerning the results obtained by our proposal, it can be seen that the auto-encoder obtains low results for this task. These are similar to those attained by the approximations based on hand-crafted features, possibly because it is trained unsupervised and learns overly generic features. Using the features learned by the networks specialized in color and shape separately, the results improve, with the best being the result obtained for the shape. In particular, the shape classifier is better than all the state-of-the-art works except \cite{Tursun2020b}. Finally, the results are further improved when our proposal combines the features in a weighted manner, surpassing the other state-of-the-art methods. When assigning more weight to shape (70\%) than to color (30\%), our method outperforms previous works by a notable margin.

\begin{table}[ht]
\setlength{\tabcolsep}{9pt}
\caption{Comparison with the previous state-of-the-art results for METU dataset. NAR is the normalized average rank metric. Smaller NAR values indicate better results.}
\label{tab:cmp_results}
\centering
\begin{tabular}{llc}
\hline
\textbf{Approach}   & \textbf{Method}                          & \textbf{NAR} \\ 
\hline
\multirow{8}{*}{Hand-crafted features} 
& Color histograms \cite{Lei1999}             & 0.400 \\
& SIFT \cite{Lowe2004}                        & 0.348 \\
& TRI-SIFT \cite{Kalantidis2011}              & 0.324 \\
& LBP \cite{Ojala2002MultiresolutionGA}       & 0.276 \\
& SURF \cite{Bay2008}                         & 0.207 \\
& OR-SIFT \cite{Kalantidis2011}               & 0.190 \\
& SIFT without text \citep{Perez2018}         & 0.154 \\
& Enhanced SIFT \cite{Feng2018}               & 0.083 \\
\hline
\multirow{6}{*}{Neural networks-based}
& GoogLeNet \cite{Szegedy2015}                                & 0.118  \\
& AlexNet \cite{Krizhevsky2012}                               & 0.112  \\
& VGG16 \cite{Simonyan2015}                                   & 0.086  \\
& Visual network \citep{Perez2018}		& 0.066	 \\
& Conceptual network \citep{Perez2018}	& 0.063	 \\
& ATRHA R-MAC \cite{Tursun2020b}                              & 0.063  \\
& Fusion of hand-crafted \& CNN features \cite{Tursun2017}    & 0.062  \\
& Fusion of visual and conceptual networks \citep{Perez2018}  & 0.047	 \\
& ATRHA CAMSA MAC \cite{Tursun2020b}                          & 0.040  \\
\hline
\multirow{5}{*}{\textbf{Our approach}} 
& Autoencoder                                                 & 0.118	\\
& Color	& 0.090	\\
& Shape	& 0.044 \\
& Weighted features (70\% color, 30\% shape)                  & 0.034 \\
& Weighted features (30\% color, 70\% shape)                  & \textbf{0.018} \\
\hline
\end{tabular}
\end{table}

\begin{figure}[ht]
\centering
		\includegraphics[width=1\linewidth]{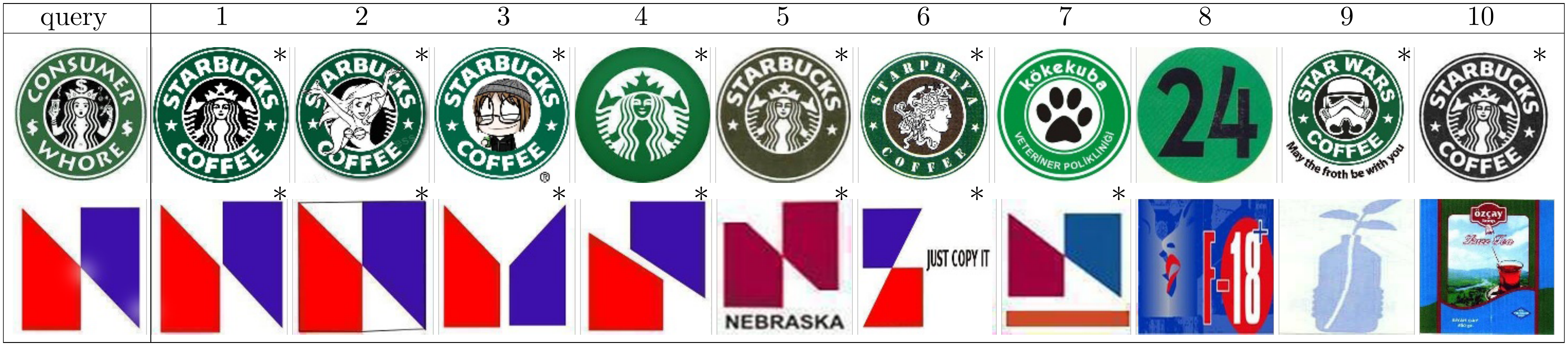}
		\caption{Two examples of the 10-nearest neighbors obtained in the METU dataset by assigning 30\% of weight to color and 70\% to shape. The first logo is the query. The correct results found are marked with an asterisk (*).}
	\label{fig:metu_results}
\end{figure}

Figure \ref{fig:metu_results} shows an example of the results obtained for the METU dataset when using our proposal combining the characteristics of shape (70\%) and color (30\%). In this figure, the first logo is the query, and the rest are the ten most similar logos ---an asterisk (*) marks the correct results. For the query in the first row, the ground truth contains thirteen similar logos. Our method found eight among the first ten results; the others are in positions 11, 16, 17, 20, and 23. For the query in the second row, the ground truth had nine similar logos. In this case, the method returned seven of them among the first ten results, and the other two were in positions 57 and 65 (out of a total of 923,340 possible logos).

\subsection{Surveys}

Since the classification of brands can often be subjective, to assess the effectiveness of our proposal, we also evaluated (using the same metrics) the results that experts in this task would obtain. 

This was achieved by surveying 107 graphic design students and professionals. In this survey, 3 logos with color labels, 3 with shapes, and 6 with figurative elements were randomly selected for each participant, asking 12 questions per participant. A reduced set of possible answers to each question was provided, and the participants were asked to mark only the labels they considered to be present in the logo.

In the color questions, the participants were shown the following statement: ``\textit{Indicate whether you can see the following colors in this logo (the white background is not considered to be a color).}''. The following 13 possible colors were then provided, and the respondent had to choose one or more: Red, Yellow, Green, Blue, Violet, White, Brown, Black, Gray, Silver, Gold, Orange, and Pink.

In the case of shape, the respondents were instructed to select the distinctive shapes when provided with 8 possible options:
1) Circles or ellipses;
2) Segments or sectors of circles or ellipses; 
3) Triangles, lines forming an angle;
4) Quadrilaterals;
5) Other polygons or geometrical figures;
6) Different geometrical figures, juxtaposed or joined;
7) Lines, bands; 
and 
8) Geometrical solids (3D objects: spheres, cubes, cylinders, pyramids, etc.).

In the case of figurative elements, since there are 123 possible labels, only the correct answers, along with another 4 or 5 incorrect answers, were given to the respondents rather than all the options.

Table \ref{tab:quiz_results} depicts the results obtained from the surveys using the same LRAP metric considered previously. These results are compared with those obtained using the CNN networks specialized in classifying these same characteristics (previously shown in Table \ref{tab:sim_results}). As can be seen, the proposed methodology improves the precision of the labeling carried out by the professionals and design students surveyed, especially in the case of the labeling of figurative elements. These results confirm the difficulty of this task owing to subjectivity when interpreting the meaning of the elements that appear in a logo or the characteristics that could be considered representative of the brand.

\begin{table}[t]
\setlength{\tabcolsep}{9pt}
\caption{Results obtained in the survey of design students and professionals using the LRAP metric, compared with the result obtained by our proposal. Higher LRAP values indicate better results.}
\label{tab:quiz_results}
\centering
\begin{footnotesize}
\begin{tabular}{lcc}
\hline
\textbf{Criteria}     
& \begin{tabular}[c]{@{}c@{}}\textbf{Students and}\\\textbf{professionals of design}\end{tabular}
& \textbf{Our proposal} \\ 
\hline
\textbf{Color}		    & 0.6735		& 0.7070     \\
\textbf{Shape}	        & 0.5467		& 0.6579     \\
\textbf{Sub-category}   & 0.3673		& 0.6850     \\
\hline
\textbf{Average}        & 0.5292		& 0.6833     \\
\hline
\end{tabular}
\end{footnotesize}
\end{table}

\subsubsection{Analysis of the survey responses}

\begin{figure}[!ht]
\begin{footnotesize}
\begin{minipage}{.49\textwidth}
\begin{tabular}{lll} 
\multicolumn{3}{c}{
\subfloat{
	\includegraphics[width=.35\textwidth]{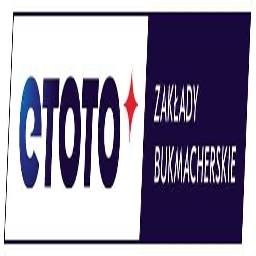}
    \label{fig:survey-color-1}
}} \\
\multicolumn{3}{l}{(a) Color labels: White; Black} \\
\hline
Answers & \# & \% \\
\hline
Two & 0 & 0  \\
One & 0 & 0 \\
Two (and others)  & 3 & 11.54  \\
One (and others)  & 7  & 26.92  \\
Others (red and/or blue) & 16 & 61.54\\
\end{tabular}
\end{minipage}
\begin{minipage}{.49\textwidth}
\begin{tabular}{lll}
\multicolumn{3}{c}{
\subfloat{
	\includegraphics[width=.35\textwidth]{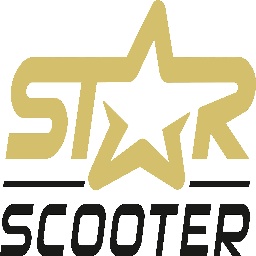}
\label{fig:survey-color-3}
}}\\
\multicolumn{3}{l}{(b) Color labels: Black; Gold} \\
\hline    
Answers & \# & \% \\
\hline
Two & 19 & 67.86 \\
One & 0 & 0 \\
Two (and others) & 2 & 7.14 \\
One (and others) & 7  & 25.00 \\
Others & 0 & 0 \\
\end{tabular}
\end{minipage}
\begin{minipage}{.49\textwidth}
\begin{tabular}{lll}
\multicolumn{3}{c}{
\subfloat{
	\includegraphics[width=.35\textwidth]{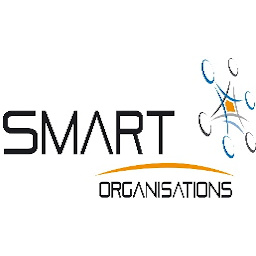}
\label{fig:survey-fig-2}
}}\\
\multicolumn{3}{l}{(c) Figurative labels: Stars, Comets;} \\
\multicolumn{3}{l}{Armillary Spheres, Planetaria, ...} \\
%  \multicolumn{3}{l}{Orbits, Atomic Models, Molecular Models.} \\
\hline    
Answers & \# & \% \\
\hline
Two & 7 & 33.33 \\
One & 8 & 38.10 \\
Two (and others) & 2 & 9.52 \\
One (and others) & 3 & 14.29 \\
Others & 0 & 0 \\
None & 1 & 4.76 \\
\end{tabular}
\end{minipage}
\begin{minipage}{.49\textwidth}
\begin{tabular}{lll}
\multicolumn{3}{c}{
\subfloat{
	\includegraphics[width=.3\textwidth]{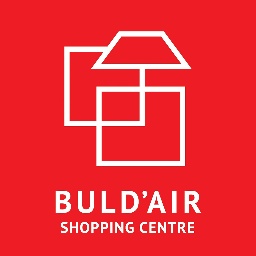}
\label{fig:survey-fig-3}
}}\\
\multicolumn{3}{l}{(d) Figurative labels: Lighting} \\
\multicolumn{3}{l}{Wireless Valves} \\
\hline    
Answers & \# & \% \\
\hline
One & 2 & 8.00 \\
One (and others) & 7 & 28.00 \\
Others & 14 & 56.00 \\
None & 2 & 8.00 \\
\end{tabular}
\end{minipage}
\begin{minipage}{.49\textwidth}
\begin{tabular}{lll}
\multicolumn{3}{c}{
\subfloat{
	\includegraphics[width=.35\textwidth]{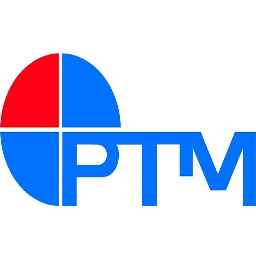}
\label{fig:survey-shape-1}
}}\\
\multicolumn{3}{l}{(e) Shape labels: Circles or ellipses; } \\
\multicolumn{3}{l}{Segments or sectors of circles or ellipses;} \\
\multicolumn{3}{l}{Lines, bands.} \\
\hline    
Answers & \# & \% \\
\hline
Three & 0 & 0 \\
Two & 6 & 15.38 \\
One & 4 & 10.26 \\
Three (and others) & 5 & 12.82\\
Two (and others) &  13 & 33.33 \\
One (and others) & 8 & 20.51 \\
Others & 3 & 7.69 \\
\end{tabular}
\end{minipage}
\begin{minipage}{.49\textwidth}
\begin{tabular}{lll}
\multicolumn{3}{c}{
\subfloat{
	\includegraphics[width=.35\textwidth]{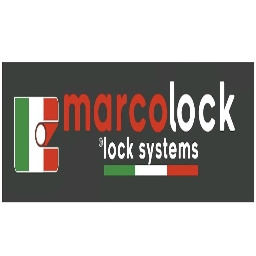}
\label{fig:survey-shape-3}
}}\\
\multicolumn{3}{l}{(f) Shape labels: Circles or ellipses; } \\
\multicolumn{3}{l}{Quadrilaterals; Geometrical solids.} \\
\hline    
Answers & \# & \% \\
\hline
Three & 0 & 0 \\
Two & 1 & 3.57 \\
One & 1 & 3.57 \\
Three (and others) & 3 & 10.71\\
Two (and others) &  9 & 32.14 \\
One (and others) & 7 & 25.00 \\
Others & 7 & 25.00 \\
\end{tabular}
\end{minipage}
\end{footnotesize}
\caption{Examples of the questions and answers in the surveys. The correct responses and a summary of the answers provided for each option are included for each question.}
\label{fig:survey_results}
\end{figure}

Figure \ref{fig:survey_results} shows some examples of the questions asked in the survey, including the correct responses (based on the database labeling) and statistical data on the number of correct answers provided by the participants. For example, if there are two possible options, the number of participants who got both or only one correct is indicated. The cases in which, in addition to having one or two correct answers, they also answered other incorrect options are also detailed.

Figures \ref{fig:survey-color-1} and \ref{fig:survey-color-3} show two examples of the color questions. In the case of the first, no respondent attained the correct answers. Most appreciated blue and red in the image, although the image was not labeled blue but black, and the color red was not labeled in the dataset. In Figure \ref{fig:survey-color-3}, most respondents selected the correct answer (some included other options), and 7 confused Yellow with Gold or Brown. As can be seen in these examples, people can appreciate color differently, either by the nature of the individual, the tone assigned to the color, or defects in the image related to the means of production. Another error source is subjectivity in the labeling process since sometimes only the color considered representative of the brand is labeled.

Concerning semantic labels, in Figure \ref{fig:survey-fig-2}, 9 of the respondents answered correctly to both questions, selecting an additional label in two cases. When analyzed individually, 95\% of the respondents recognized one of the two labels. On the other hand, in Figure \ref{fig:survey-fig-3}, which is labeled with a single class, only 36\% marked the correct answer, with the majority selecting other options such as ``Furniture'', ``Electrical Equipment'' or ``Heating, Cooking Or Refrigerating Equipment, Washing Machines, Drying Equipment''. As will be noted, recognizing semantic elements in a logo is not a trivial task. In many cases, figures are oversimplified and may be confused with other representations. In addition, the interpretation often depends on the individuals who perceive it and their cultural, personal, or professional background.

Figure \ref{fig:survey-shape-1} is labeled with three shape classes. Of these, 12 of the respondents (out of a total of 39) got only one correct. 
%(8 of them included incorrect answers), 19 detected two (most of them included wrong responses), and only 5 selected the three correct ones (all included other wrong answers). 
The case of Figure \ref{fig:survey-shape-3} is similar since the answers are multiple and there are very different combinations. In this case, the image contains ``Geometrical solids (3D objects)'', and only 5 of the 28 people who evaluated this logo marked this answer. These examples illustrate the complexity of detecting all the shapes in an image. It is usually somewhat subjective if a shape is representative of a logo design. Moreover, a predominant shape can sometimes influence the observer to ignore other shapes in the image.

% ---------------------------------------------------------------------------
\section{Conclusions}
\label{sec:conclusions}

This paper presents a methodology for the multi-label classification of logos, considering main characteristics such as color, shape, semantic elements, and text. Furthermore, the proposed method also allows obtaining a ranking of the most similar logos, in which users can select the characteristics to consider in the search process. To the best of our knowledge, no other methods in the literature address these two objectives. Therefore, a proposal of this kind is of great interest, both methodologically and practically, as regards assisting in multiple tasks, such as labeling logos, detecting plagiarism, or similarities between brands.

The proposed architecture combines, in a weighted fashion, the representation learned by a series of multi-label classification networks that specialize in detecting the most distinctive characteristics of logos. Moreover, the method performs a preprocessing stage to remove uniform backgrounds and text from input images. The experiments showed that removing the text from the logo helps classify the shape, but not other types of characteristics. This may be because the text often includes representative characteristics of the logo, such as color or figurative elements, and removing them worsens the result.

The experimental results show that the proposed approach is reliable for both classification and similarity search. Furthermore, the comparison made with 17 state-of-the-art TIR methods shows that our proposal is notably better than previous approaches, especially considering color and shape. 

This paper also studies the logo labeling issues in trademark registration databases since only the most distinctive characteristics of the brand are generally labeled by registration agencies, resulting in incomplete and often inconsistent labeling. Moreover, the semantics of trademarks can be subjective, which results in difficulties for operators. These problems are produced either by the labeling process itself or are motivated by the Vienna coding since it is a closed categorization and some characteristics are challenging to define. 

One of the proposed methodology's advantages is aiding in this task, since it suggests an initial classification that follows homogeneous criteria, which, in addition to facilitating the work, is complete and exhaustive. Furthermore, given that many people label ground-truth data, an automatic classification method reduces the inconsistency of human subjectivity caused by the different perceptions of the same visual representation and the difficulty of expressing graphic qualities in words.

We also performed a qualitative evaluation, which was carried out with expert designers to assess labeling consistency. These experiments showed that the proposed methodology provides better labeling than a human operator would assign, even in the case of experts in this task. The labeling suggested by the system could be used as an initial proposal to be reviewed by the operator. In addition, students and design professionals could use the system as aid since they could check the labeling proposal for a new design, search for references, ideas, and styles, or detect similar marks and possible plagiarism.

%---------------------------
\subsection*{Acknowledgments}

This work is supported by the Pattern Recognition and Artificial Intelligence Group (GRFIA) from the University of Alicante and the University Institute for Computing Research (IUII). Some of the computing resources used in this project are provided by the Valencian Government and FEDER through IDIFEDER/2020/003.

%---------------------------
%\bibliographystyle{model5-names}
%\biboptions{authoryear}

\bibliographystyle{elsarticle-num}

\bibliography{main}

\begin{thebibliography}{10}
\expandafter\ifx\csname url\endcsname\relax
  \def\url#1{\texttt{#1}}\fi
\expandafter\ifx\csname urlprefix\endcsname\relax\def\urlprefix{URL }\fi
\expandafter\ifx\csname href\endcsname\relax
  \def\href#1#2{#2} \def\path#1{#1}\fi

\bibitem{Bianco_2017}
S.~Bianco, M.~Buzzelli, D.~Mazzini, R.~Schettini, Deep learning for logo
  recognition, Neurocomputing 245 (2017) 23–30.
\newblock \href {https://doi.org/10.1016/j.neucom.2017.03.051}
  {\path{doi:10.1016/j.neucom.2017.03.051}}.

\bibitem{10.1007/978-3-030-20518-8_11}
O.~Orti, R.~Tous, M.~Gomez, J.~Poveda, L.~Cruz, O.~Wust, Real-time logo
  detection in brand-related social media images, in: I.~Rojas, G.~Joya,
  A.~Catala (Eds.), Advances in Computational Intelligence, Springer
  International Publishing, Cham, 2019, pp. 125--136.

\bibitem{Kostinger_planartrademark}
M.~Köstinger, P.~M. Roth, H.~Bischof, Planar trademark and logo retrieval,
  Tech. rep., Computer Graphics and Vision, Graz University of Technology,
  Austria (2010).

\bibitem{Perez2018}
C.~A. {Perez}, P.~A. {Estévez}, F.~J. {Galdames}, D.~A. {Schulz}, J.~P.
  {Perez}, D.~{Bastías}, D.~R. {Vilar}, {Trademark Image Retrieval Using a
  Combination of Deep Convolutional Neural Networks}, in: Int. Joint Conference
  on Neural Networks (IJCNN), 2018, pp. 1--7.

\bibitem{Schietse2007}
J.~Schietse, J.~Eakins, R.~Veltkamp, Practice and challenges in trademark image
  retrieval, in: Proceedings of the 6th ACM International Conference on Image
  and Video Retrieval, CIVR 2007, 2007, pp. 518--524.
\newblock \href {https://doi.org/10.1145/1282280.1282355}
  {\path{doi:10.1145/1282280.1282355}}.

\bibitem{Logos01}
Capsule, Design Matters: Logos 01: An Essential Primer for Today's Competitive
  Market, Rockport Publishers, 2007.

\bibitem{ibpriaLogos}
A.-J. Gallego, A.~Pertusa, M.~Bernabeu, Multi-label logo classification using
  convolutional neural networks, in: A.~Morales, J.~Fierrez, J.~S. S{\'a}nchez,
  B.~Ribeiro (Eds.), Pattern Recognition and Image Analysis, Springer
  International Publishing, Cham, 2019, pp. 485--497.

\bibitem{DudaHS01}
R.~O. Duda, P.~E. Hart, D.~G. Stork, Pattern classification, 2nd Edition,
  Wiley, 2001.

\bibitem{Ghosh2015}
S.~Ghosh, R.~Parekh, Automated color logo recognition system based on shape and
  color features, Int. Journal of Computer Applications 118~(12) (2015) 13--20.

\bibitem{ShapeTrademark10}
H.~Qi, K.~Li, Y.~Shen, W.~Qu, An effective solution for trademark image
  retrieval by combining shape description and feature matching, Pattern
  Recognition 43~(6) (2010) 2017--2027.

\bibitem{chiambrand}
J.-H. Chiam, Brand logo classification, Tech. rep., Stanford University (2015).

\bibitem{KUMAR2016370}
N.~V. Kumar, Pratheek, V.~V. Kantha, K.~Govindaraju, D.~Guru, Features fusion
  for classification of logos, in: Int. Conf. on Computational Modelling and
  Security (CMS), Vol.~85, 2016, pp. 370--379.

\bibitem{Guru18}
D.~S. Guru, N.~Vinay~Kumar, {Interval Valued Feature Selection for
  Classification of Logo Images}, in: A.~Abraham, P.~K. Muhuri, A.~K. Muda,
  N.~Gandhi (Eds.), Intelligent Systems Design and Applications, Springer
  International Publishing, Cham, 2018, pp. 154--165.

\bibitem{IandolaSGK15}
F.~N. Iandola, A.~Shen, P.~Gao, K.~Keutzer, {DeepLogo: Hitting Logo Recognition
  with the Deep Neural Network Hammer}, CoRR abs/1510.02131 (2015).
\newblock \href {http://arxiv.org/abs/1510.02131} {\path{arXiv:1510.02131}}.

\bibitem{Pornpanomchai2015}
C.~Pornpanomchai, P.~Boonsripornchai, P.~Puttong, C.~Rattananirundorn, Logo
  recognition system, in: 2015 International Computer Science and Engineering
  Conference (ICSEC), IEEE, 2015, pp. 1--6.

\bibitem{lourenco2019hierarchyofvisualwords}
V.~N. Lourenço, G.~G. Silva, L.~A.~F. Fernandes, Hierarchy-of-visual-words: a
  learning-based approach for trademark image retrieval (2019).
\newblock \href {http://arxiv.org/abs/1908.02786} {\path{arXiv:1908.02786}}.

\bibitem{lecun2015deep}
Y.~LeCun, Y.~Bengio, G.~Hinton, Deep learning, Nature 521~(7553) (2015)
  436--444.

\bibitem{Krizhevsky2012}
A.~Krizhevsky, I.~Sutskever, G.~E. Hinton, Imagenet classification with deep
  convolutional neural networks, in: F.~Pereira, C.~J.~C. Burges, L.~Bottou,
  K.~Q. Weinberger (Eds.), Advances in Neural Information Processing Systems,
  Vol.~25, Curran Associates, Inc., 2012.

\bibitem{Xia2019}
Z.~Xia, J.~Lin, X.~Feng, Trademark image retrieval via transformation-invariant
  deep hashing, Journal of Visual Communication and Image Representation 59
  (2019) 108--116.
\newblock \href {https://doi.org/https://doi.org/10.1016/j.jvcir.2019.01.011}
  {\path{doi:https://doi.org/10.1016/j.jvcir.2019.01.011}}.

\bibitem{ZhangZhou2014}
M.-L. Zhang, Z.-H. Zhou, A review on multi-label learning algorithms, IEEE
  Trans. on Knowledge and Data Engineering 26 (2014) 1819--1837.
\newblock \href {https://doi.org/10.1109/TKDE.2013.39}
  {\path{doi:10.1109/TKDE.2013.39}}.

\bibitem{MLCtext}
H.~Dong, W.~Wang, K.~Huang, F.~Coenen, Automated social text annotation with
  joint multi-label attention networks, IEEE Trans. on Neural Networks and
  Learning Systems 32~(5) (2020) 2224--2238.
\newblock \href {https://doi.org/10.1109/TNNLS.2020.3002798}
  {\path{doi:10.1109/TNNLS.2020.3002798}}.

\bibitem{MLCmusic}
K.~Trohidis, G.~Tsoumakas, G.~Kalliris, I.~Vlahavas, Multi-label classification
  of music by emotion, EURASIP Journal on Audio, Speech, and Music Processing
  2011~(1) (2011) 1--9.
\newblock \href {https://doi.org/10.1186/1687-4722-2011-426793}
  {\path{doi:10.1186/1687-4722-2011-426793}}.

\bibitem{MLCscene}
M.~R. Boutell, J.~Luo, X.~Shen, C.~M. Brown, Learning multi-label scene
  classification, Pattern Recognition, 37(9) (2004) 1757--1771.

\bibitem{Kalantidis2011}
Y.~Kalantidis, L.~G. Pueyo, M.~Trevisiol, R.~van Zwol, Y.~Avrithis, Scalable
  triangulation-based logo recognition, in: Proceedings of the 1st ACM
  International Conference on Multimedia Retrieval, ICMR '11, Association for
  Computing Machinery, New York, NY, USA, 2011.
\newblock \href {https://doi.org/10.1145/1991996.1992016}
  {\path{doi:10.1145/1991996.1992016}}.

\bibitem{Sage_2018}
A.~Sage, E.~Agustsson, R.~Timofte, L.~Van~Gool, Logo synthesis and manipulation
  with clustered generative adversarial networks, in: Proceedings of the IEEE
  Conference on Computer Vision and Pattern Recognition, 2018, pp. 5879--5888.
\newblock \href {https://doi.org/10.1109/cvpr.2018.00616}
  {\path{doi:10.1109/cvpr.2018.00616}}.

\bibitem{Tursun2015METU}
O.~Tursun, S.~Kalkan, {METU dataset: A big dataset for benchmarking trademark
  retrieval}, in: 2015 14th IAPR International Conference on Machine Vision
  Applications (MVA), IEEE, 2015, pp. 514--517.

\bibitem{Tursun2017}
O.~Tursun, C.~Aker, S.~Kalkan, A large-scale dataset and benchmark for similar
  trademark retrieval, CoRR abs/1701.05766 (2017).
\newblock \href {http://arxiv.org/abs/1701.05766} {\path{arXiv:1701.05766}}.

\bibitem{Tzk2018OpenSL}
A.~T{\"u}zk{\"o}, C.~Herrmann, D.~Manger, J.~Beyerer, Open set logo detection
  and retrieval, in: Int. Joint Conf. on Computer Vision, Imaging and Computer
  Graphics Theory and Applications (VISIGRAPP), 2018.

\bibitem{world2002international}
{World Intellectual Property Organization}, International Classification of the
  Figurative Elements of Marks: (Vienna Classification)., WIPO publication,
  World Intellectual Property Organization, 2002.

\bibitem{Rusinol2011}
M.~Rusi{\~{n}}ol, D.~Aldavert, D.~Karatzas, R.~Toledo, J.~Llad{\'o}s,
  Interactive trademark image retrieval by fusing semantic and visual content,
  in: European Conference on Information Retrieval, Springer, 2011, pp.
  314--325.

\bibitem{wheeler2013designing}
A.~Wheeler, Designing Brand Identity: An Essential Guide for the Whole Branding
  Team, Wiley, 2013.

\bibitem{chaves2003marca}
N.~Chaves, R.~Belluccia, La Marca Corporativa: Gesti{\'o}n y Dise{\~n}o de
  S{\'\i}mbolos y Logotipos, Estudios de Comunicaci{\'o}n Series, Paid{\'o}s,
  2003.

\bibitem{baek2019character}
Y.~Baek, B.~Lee, D.~Han, S.~Yun, H.~Lee, Character region awareness for text
  detection, in: Proceedings of the IEEE Conference on Computer Vision and
  Pattern Recognition (CVPR), 2019, pp. 9365--9374.

\bibitem{wang2018image}
Y.~Wang, X.~Tao, X.~Qi, X.~Shen, J.~Jia, Image inpainting via generative
  multi-column convolutional neural networks, in: Advances in Neural
  Information Processing Systems, 2018, pp. 331--340.
\newblock \href {https://doi.org/10.1016/j.displa.2021.102028}
  {\path{doi:10.1016/j.displa.2021.102028}}.

\bibitem{lecun2010}
Y.~LeCun, K.~Kavukcuoglu, C.~Farabet, Convolutional networks and applications
  in vision, in: Proceedings of 2010 IEEE International Symposium on Circuits
  and Systems (ISCAS), 2010, pp. 253--256.

\bibitem{BatchNormalization}
S.~Ioffe, C.~Szegedy, Batch normalization: Accelerating deep network training
  by reducing internal covariate shift, in: International Conference on Machine
  Learning (ICML), 2015, pp. 448--456.

\bibitem{Srivastava2014}
N.~Srivastava, G.~Hinton, A.~Krizhevsky, I.~Sutskever, R.~Salakhutdinov,
  Dropout: A simple way to prevent neural networks from overfitting, Journal of
  Machine Learning Research 15~(1) (2014) 1929--1958.

\bibitem{Glorot2011relu}
X.~Glorot, A.~Bordes, Y.~Bengio, {Deep Sparse Rectifier Neural Networks}, in:
  G.~Gordon, D.~Dunson, M.~Dudík (Eds.), Proceedings of the Fourteenth
  International Conference on Artificial Intelligence and Statistics, Vol.~15
  of Proceedings of Machine Learning Research, PMLR, Fort Lauderdale, FL, USA,
  2011, pp. 315--323.

\bibitem{Babenko:2014}
A.~Babenko, A.~Slesarev, A.~Chigorin, V.~Lempitsky, Neural codes for image
  retrieval, in: European Conference on Computer Vision (ECCV), Springer, 2014,
  pp. 584--599.

\bibitem{Gallego2020}
A.~J. {Gallego}, J.~{Calvo-Zaragoza}, J.~R. {Rico-Juan}, Insights into
  efficient k-nearest neighbor classification with convolutional neural codes,
  IEEE Access 8 (2020) 99312--99326.
\newblock \href {https://doi.org/10.1109/ACCESS.2020.2997387}
  {\path{doi:10.1109/ACCESS.2020.2997387}}.

\bibitem{Huang06}
F.~Huang, Y.~LeCun, {Large-scale learning with SVM and convolutional nets for
  generic object categorization}, in: Proceedings of the IEEE Computer Society
  Conference on Computer Vision and Pattern Recognition, CVPR 2006, Vol.~1,
  2006, pp. 284--291.

\bibitem{Razavian2014}
A.~S. Razavian, H.~Azizpour, J.~Sullivan, S.~Carlsson, {CNN Features
  Off-the-Shelf: An Astounding Baseline for Recognition}, in: Proceedings of
  the 2014 IEEE Conference on Computer Vision and Pattern Recognition
  Workshops, CVPRW '14, IEEE Computer Society, Washington, DC, USA, 2014, pp.
  512--519.

\bibitem{hinton1994autoencoders}
G.~E. Hinton, R.~S. Zemel, {Autoencoders, Minimum Description Length and
  Helmholtz Free Energy}, in: Advances in Neural Information Processing
  Systems, 1994, pp. 3--10.

\bibitem{Baldi2012autoencoders}
P.~Baldi, Autoencoders, unsupervised learning, and deep architectures, in:
  {Proceedings of ICML workshop on unsupervised and transfer learning}, 2012,
  pp. 37--49.

\bibitem{ZhengZWWT16}
L.~Zheng, Y.~Zhao, S.~Wang, J.~Wang, Q.~Tian, Good practice in {CNN} feature
  transfer abs/1604.00133 (2016).
\newblock \href {https://doi.org/10.48550/ARXIV.1604.00133}
  {\path{doi:10.48550/ARXIV.1604.00133}}.

\bibitem{Gallego2018knn}
A.-J. Gallego, A.~Pertusa, J.~Calvo-Zaragoza, Improving convolutional neural
  networks’ accuracy in noisy environments using k-nearest neighbors, Applied
  Sciences 8~(11) (2018).

\bibitem{EleftheriosSpyromitros2008}
I.~V. Eleftherios~Spyromitros, Grigorios~Tsoumakas, An empirical study of lazy
  multilabel classification algorithms, in: Proc. 5th Hellenic Conference on
  Artificial Intelligence (SETN 2008), Springer, 2008, pp. 401--406.

\bibitem{Bottou2010}
L.~Bottou, Large-scale machine learning with stochastic gradient descent, in:
  Proceedings of the 9th International Conference on Computational Statistics,
  COMPSTAT 2010, Springer, 2010, pp. 177--186.

\bibitem{Zeiler12}
M.~D. Zeiler, {ADADELTA:} an adaptive learning rate method, arXiv preprint
  arXiv:1212.5701 (2012).

\bibitem{Tsoumakas10miningmulti-label}
G.~Tsoumakas, I.~Katakis, I.~Vlahavas, Mining multi-label data, in: {Data
  Mining and Knowledge Discovery Handbook}, Springer, 2010, pp. 667--685.

\bibitem{Maaten2008}
G.~van~der Maaten, Laurens;~Hinton, {Visualizing High-Dimensional Data Using
  t-SNE}, Journal of Machine Learning Research 9~(11) (2008) 2579--2605.

\bibitem{Lei1999}
Z.~Lei, L.~Fuzong, Z.~Bo, A cbir method based on color-spatial feature, in:
  TENCON, Proceedings of the IEEE Region 10 Conference., Vol.~1, 1999, pp.
  166--169.
\newblock \href {https://doi.org/10.1109/TENCON.1999.818376}
  {\path{doi:10.1109/TENCON.1999.818376}}.

\bibitem{Ojala2002MultiresolutionGA}
T.~Ojala, M.~Pietik{\"a}inen, T.~M{\"a}enp{\"a}{\"a}, Multiresolution
  gray-scale and rotation invariant texture classification with local binary
  patterns, {IEEE Trans. on Pattern Analysis and Machine Intelligence} 24
  (2002) 971--987.

\bibitem{Lowe2004}
D.~G. Lowe, Distinctive image features from scale-invariant keypoints,
  International journal of computer vision 60~(2) (2004) 91–110.
\newblock \href {https://doi.org/10.1023/B:VISI.0000029664.99615.94}
  {\path{doi:10.1023/B:VISI.0000029664.99615.94}}.

\bibitem{Bay2008}
H.~Bay, A.~Ess, T.~Tuytelaars, L.~{Van Gool}, {Speeded-Up Robust Features
  (SURF)}, Computer Vision and Image Understanding 110~(3) (2008) 346--359,
  similarity Matching in Computer Vision and Multimedia.
\newblock \href {https://doi.org/https://doi.org/10.1016/j.cviu.2007.09.014}
  {\path{doi:https://doi.org/10.1016/j.cviu.2007.09.014}}.

\bibitem{Feng2018}
Y.~Feng, C.~Shi, C.~Qi, J.~Xu, B.~Xiao, C.~Wang, Aggregation of reversal
  invariant features from edge images for large-scale trademark retrieval, in:
  2018 4th International Conference on Control, Automation and Robotics
  (ICCAR), 2018, pp. 384--388.
\newblock \href {https://doi.org/10.1109/ICCAR.2018.8384705}
  {\path{doi:10.1109/ICCAR.2018.8384705}}.

\bibitem{Szegedy2015}
C.~Szegedy, W.~Liu, Y.~Jia, P.~Sermanet, S.~Reed, D.~Anguelov, D.~Erhan,
  V.~Vanhoucke, A.~Rabinovich, Going deeper with convolutions, in: 2015 IEEE
  Conference on Computer Vision and Pattern Recognition (CVPR), 2015, pp. 1--9.
\newblock \href {https://doi.org/10.1109/CVPR.2015.7298594}
  {\path{doi:10.1109/CVPR.2015.7298594}}.

\bibitem{Simonyan2015}
K.~Simonyan, A.~Zisserman, Very deep convolutional networks for large-scale
  image recognition, in: Y.~Bengio, Y.~LeCun (Eds.), 3rd International
  Conference on Learning Representations, {ICLR}, 2015.

\bibitem{Tursun2020b}
O.~Tursun, S.~Denman, S.~Sivapalan, S.~Sridharan, C.~Fookes, S.~Mau,
  Component-based attention for large-scale trademark retrieval, IEEE Trans. on
  Information Forensics and Security (2020).
\newblock \href {https://doi.org/10.1109/tifs.2019.2959921}
  {\path{doi:10.1109/tifs.2019.2959921}}.

\end{thebibliography}
%---------------------------

%--------------------------------------------------------------------------------------------
\appendix

\section{Vienna classification}
\label{app:vienna}

Next, we include the list of labels used in Vienna classification \cite{world2002international}. In the scope of this work, figurative elements are those with codes from 1 to 25. Codes from 26 onwards are related to shape, text and color, and were not used for figurative classification.

{\small
\begin{longtable}{|p{2cm}|p{11cm}|}
\hline
1 & Celestial bodies, Natural Phenomena, Geographical Maps. \\
2 & Human beings. \\
3 & Animals. \\
4 & Supernatural, fabulous, fantastic or unidentifiable Beings. \\
5 & Plants. \\
6 & Landscapes. \\
7 & Constructions, structures for advertisements, gates or Barriers. \\
8 & Foodstuffs. \\
9 & Textiles, clothing, sewing accessories, headwear, footwear. \\
10 & Tobacco, smokers’ requisites, matches, travel goods, fans, toilet articles. \\
11 & Household utensils. \\
12 & Furniture, sanitary installations. \\
13 & Lighting, wireless valves, heating, cooking or refrigerating equipment, washing machines, drying equipment. \\
14 & Ironmongery, tools, ladders. \\
15 & Machinery, motors, engines \\
16 & Telecommunications, sound recording or reproduction, computers, photography, cinematography, optics. \\
17 & Horological instruments, jewelry, weights and measures. \\
18 & Transport, equipment for animals. \\
19 & Containers and packing, representations of miscellaneous products. \\
20 & Writing, drawing or painting materials, office requisites, stationery and booksellers' goods. \\
21 & Games, toys, sporting articles, roundabouts. \\
22 & Musical instruments and their accessories, music accessories, bells, pictures, sculptures. \\
23 & Arms, ammunition, armour. \\
24 & Heraldry, coins, emblems, symbols. \\
25 & Ornamental motifs, surfaces or backgrounds with ornaments. \\
\hline
26 & Geometrical figures and solids. \\
27 & Forms of writing, numerals. \\
28 & Inscriptions in various characters. \\
29 & Colors. \\
\hline
\end{longtable}
}

Vienna codes used in this work for colors (code 29 of the main Vienna classification).

{\small
\begin{longtable}{|p{2cm}|p{11cm}|}
\hline
29.01.01 & Red. \\
29.01.02 & Yellow. \\
29.01.03 & Green. \\
29.01.04 & Blue. \\
29.01.05 & Violet. \\
29.01.06 & White. \\
29.01.07 & Brown. \\
29.01.08 & Black. \\
29.01.95 & Silver. \\
29.01.96 & Gray. \\
29.01.97 & Gold. \\
29.01.98 & Orange. \\
29.01.99 & Pink. \\
\hline
\end{longtable}
}

Vienna codes used in this work for shapes (code 26 of the main Vienna classification).

{\small
\begin{longtable}{|p{1cm}|p{12cm}|}
\hline
26.1 & Circles, ellipses. \\
26.2 & Segments or sectors of circles or ellipses. \\
26.3 & Triangles, lines forming an angle. \\
26.4 & Quadrilaterals. \\
26.5 & Other polygons. This category also groups 26.13 (Other geometrical figures, indefinable designs) and 26.7 (Different geometrical figures, juxtaposed, joined, or intersecting). \\
26.11 & Lines, bands. \\
26.15 & Geometrical solids. \\
\hline
\end{longtable}
}

\end{document}